\documentclass[conference]{IEEEtran}
\IEEEoverridecommandlockouts
\usepackage{cite}
\usepackage{amsmath,amssymb,amsfonts}
\usepackage{algorithmic}
\usepackage{graphicx}
\usepackage{textcomp}
\usepackage{xcolor,soul}
\usepackage{comment,balance}
\usepackage{subcaption,enumitem}
\usepackage{tabularx,booktabs}
\usepackage{color,colortbl}
\def\BibTeX{{\rm B\kern-.05em{\sc i\kern-.025em b}\kern-.08em
    T\kern-.1667em\lower.7ex\hbox{E}\kern-.125emX}}
    
\definecolor{Gray}{gray}{0.92}

\begin{document}

\title{Learning from Power Signals: An Automated Approach to Electrical Disturbance Identification Within a Power Transmission System\\
\thanks{This work is supported by funding through the Tennessee Valley Authority (TVA).}
}
\author{\IEEEauthorblockN{Jonathan D. Boyd$^{1\ast}$, Joshua H. Tyler$^{1\dagger}$, Anthony M. Murphy$^{2}$, Donald R. Reising$^{1}$}
\IEEEauthorblockA{$^{1}$\textit{The University of Tennessee at Chattanooga, Chattanooga, TN, USA} \\
\textit{Email: $^{\ast}$yms773@mocs.utc.edu, $^{\dagger}$ygm111@mocs.utc.edu, donald-reising@utc.edu}\\
$^{2}$\textit{Tennessee Valley Authority, Chattanooga, TN, USA}\\
\textit{Email: ammurphy@tva.gov}
}
}
\maketitle

\begin{abstract}
As power quality becomes a higher priority in the electric utility industry, the amount of disturbance event data continues to grow. Utilities simply do not have the required personnel to analyze each event by-hand. This work presents an automated approach for the analysis of power quality events recorded by digital fault recorders and power quality monitors operating within a power transmission system. The automated approach leverages rule-based analytics to examine the time and frequency domain characteristics of the voltage and current signals, and customizable thresholds are set to categorize each disturbance event. The events analyzed within this work include: various faults, motor starting, and incipient instrument transformer failure. Analytics for fourteen different event types have been developed. The analytics were tested on 160 signal files and yielded an average accuracy of 99\%. Continuous, nominal signal data analysis is performed using an approach coined as the cyclic histogram. The cyclic histogram process will be integrated into the digital fault recorders themselves to facilitate detection of subtle signal variations that are too small to trigger a disturbance event and that can occur over the course of hours or days. In addition to reducing memory requirements by a factor of 320, it is anticipated that cyclic histogram processing will aid in identification of incipient events and identifiers. This project is expected to save engineers time by automating the classification of disturbance events as well as 
 increase the reliability of the transmission system by providing near real--time detection and identification of disturbances as well as prevention of problems before they occur.
\end{abstract}


\section{Introduction}
The continued and increasing deployment of ``smart'' devices (e.g., switches, relays, etc.) within power utility generation, transmission, and distribution infrastructure has led to the recording and storage of an ever-growing amount of event data. Processing and analysis of this event data has been traditionally conducted by power utility personnel using ``by-hand'' approaches. By-hand approaches rely heavily upon the knowledge, experience, and expertise of the person or persons conducting the analysis and severely limits the number of events that can be analyzed within a given period of time. These limitations are exacerbated when considering that: (i) power utilities are unable to dedicate personnel solely to the task of event processing and analysis as well as (ii) that analysis is often conducted hours if not days after the event has occurred, thus limiting its value.

The work in~\cite{bib:Rodrig} details a rule-based approach for categorizing Power Quality (PQ) events using the S Transform (ST). The data used in this approach is a mix of simulated data and real-world data from the power system. The Fourier Transform (FT) and the Short-Time Fourier Transform (STFT) have not proven to be effective in extracting unique features of each signal. The Wavelet Transform (WT) has been used as it can extract time and frequency domain characteristics simultaneously, but it is also somewhat vulnerable to noise and computationally expensive. The ST can be thought of as a hybrid between the STFT and WT since it has the time and frequency domain characteristics, but it also uses a variable window length to provide information at different resolutions. The ST has been shown to provide better noise immunity. 
Finally, categorization of the PQ events was performed using 
Artificial Neural Networks (ANNs), fuzzy logic, decision trees, and others. The ST contours 
highlight the distinctive features present within the original PQ event signal, such as a voltage sag. A set of rules is then defined to set the thresholds needed to trigger certain event types. These rules rely heavily upon the knowledge of PQ experts and a data set containing distorted signals is used to determine the corresponding threshold values. 
The rules are designed to separate the events into three categories: magnitude disturbances, transients, and signal distortion. The tests performed on the signals include positive tests and negative tests for an extra layer of classification. This approach is also very portable to other applications due the normalization of the voltage to one to facilitate use of any voltage level. 
The results of the work in~\cite{bib:Rodrig} heavily favor the rule-based ST approach. This approach classified the disturbances with 98\% accuracy while a traditional ANN method achieved an accuracy of 92\%.  
The rule-based method can also withstand a considerable level of noise in the signal. One reason for this superior accuracy is that the rule-based approach is more specialized to each type of disturbance than the ANN approach.

The approach in~\cite{bib:Kapoor} used a machine learning approach that is augmented through the inclusion of 
the Kullback–Leibler (KL) divergence measure and standard deviation. The KL divergence is very efficient as it can be applied to a single cycle of the signal. 
The KL divergence calculates the probabilities that a particular cycle is a member of two or more events. Standard deviation is also used as it is very effective in the detection of PQ disturbances. These two methods are used for each cycle of the disturbed signal and compared with an ideal sinusoidal signal to capture the disturbance. After the detection phase, the classification phase is performed using a 
Support Vector Machine (SVM) to determine a decision boundary between event types. This method proved very effective in differentiating between events such as voltage sag and swell. However, voltage flicker and swell are more similar than sag and swell, so this approach likely will not function as well. Overall, this method achieved an accuracy of 94.02\%.

The approach in~\cite{bib:De} provides a novel PQ disturbance classification method. 
The method extracts features from 
the cross-correlogram of the PQ disturbances. The positive peak and two adjacent negative peaks are used as the classification features. 
Those three values are then fed into a fuzzy-based classification system. One drawback to the work in~\cite{bib:De} is its use of simulated data that was generated using MATLAB\textsuperscript{\textregistered}, thus classification accuracy may change when real-world data is used. The two types of correlation are cross-correlation and auto-correlation. Cross-correlation measures the strength of similarity between two signals, while auto-correlation is the cross-correlation of a signal with itself. The work in~\cite{bib:De} calculates the cross-correlation response between an ideal signal with a disturbed one to detect the disturbance. 
A fuzzy logic classifier is used to allow 
for uncertainty in a logic system. The rules in the fuzzy system are designed by human experts, so the system is only as good as those who designed it. The system used in this approach is the Mamdani-type inference system with three inputs and one output. Eighteen linguistic variables are used for the output membership function to determine the PQ event classification. This classifier was tested using seventy MATLAB\textsuperscript{\textregistered} generated signals 
and achieved an accuracy of 100\%. The accuracy remained 100\% even when noise was added to the test signals.

The work presented herein uses a series of algorithms developed in MATLAB\textsuperscript{\textregistered} R2020b that classify various PQ events into one or more categories. 
The developed algorithms are rule-based in nature with customizable thresholds based on engineers' expertise. Each PQ event's signal data is stored in a Comma Separated Values (CSV) file--generated by the field device--containing: a time vector, three voltage phases, and three current phases. A MATLAB\textsuperscript{\textregistered} executable is initiated to read each CSV file into a working directory then categorize them as a particular PQ event type or types. The latter accounts for the case of multiple PQ event types occurring and being recorded within the same CSV file. A CSV file is then generated with the classification results as well as analytic outputs such as current magnitude. Below are several differentiating factors that make the presented work unique and preferable to other methods:

\begin{itemize}[leftmargin=*]
    \item The automated process was developed and tested using real-world data rather than simulated data. All data was recorded 
    by smart field devices--PQ monitors and Digital Fault Recorders (DFRs)--operating in a high-voltage transmission system.
    \item The rule-based methods mimic the expertise of an engineer in an effort to ease interpretation and understanding of the classification results by power system personnel. 
    \item The developed algorithms contain very few MATLAB\textsuperscript{\textregistered} specific functions. 
    This reduces the need for expensive MATLAB\textsuperscript{\textregistered} Toolbox licenses 
    while allowing the algorithms to be translated into 
    other programming languages and software based upon the specific needs of the power utility. This approach is adopted to facilitate widespread use of the developed algorithms across the power industry. 
    \item The rule-based nature of the developed process allows every threshold to be changed as needed by power utility personnel based on performance or system specifics. In this paper, empirical thresholds are designated as $\tau$ in equations and as \textbf{bold} lettering in sentences.
    \item The methods used are very detailed and will predict the actual disturbance (e.g., ferroresonance) that occurred on the power system rather than simple signal characteristics like voltage sag and swell.
\end{itemize}

Another aspect of the project was to analyze continuous oscillography data that is stored on the DFRs. Each day of data can be as much as twenty to fifty gigabytes (GB), which is far too much data for an engineer to analyze manually. Due to on-board memory constraints, each DFR stores two weeks of continuous oscillography data before it is overwritten. The approach in this work uses a method known as a cyclic histogram~\cite{bib:cooke} to reduce an average day's thirty-five~{GB} worth of continuous oscillography data down to seventy-two megabytes (MB). This memory reduction not only increases the time window of how long the data can be stored--from two to roughly 1,000 weeks--but also allows 
engineers to monitor for trends and subtle deviations in continuous signal data that has not produced a disturbance large enough to trigger a DFR event.

The remainder of this paper is as follows. Section~\ref{sec:methodology} presents the methodology including general calculations, continuous waveform analysis, and the various disturbance event types. Section~\ref{sec:results} provides the results of each event type and continuous waveform analysis. Section~\ref{sec:conclusion} provides a summary and lists some opportunities for future work.

\section{Methodology}\label{sec:methodology}

This section first presents descriptions of calculation, analyses, and tests that are used in the categorization of multiple events. A specific event may require the threshold of one or more of these general calculations, analyses, or tests to be changed and are detailed under the specific event being categorized. The remainder of this section describes the methodologies developed and employed for the categorization of specific events and continuous signal processing using the cyclic histogram. 

\subsection{General Calculations, Analyses, and Tests}
\subsubsection{Calculating Nominal Values} \label{sec:nominal_values}

The first task in the processing of a voltage or current signal is to calculate nominal values from the data itself. The sampling frequency is calculated by,
\begin{equation}
    F_s=\dfrac{N}{t_{e}-t_1},
    \label{eq:sample_rate}
\end{equation}
where $F_s$ is the sampling frequency in Hertz (Hz), $N$ is the number of samples in the time vector, and $t_1$ and $t_{e}$ are the first and last values of the time vector, respectively. After the sampling frequency is known, the nominal number of samples in each cycle is determined by,
\begin{equation}
    N_{c} = \dfrac{F_s}{F_{n}},
    \label{eq:spc}
\end{equation}
where $N_{c}$ represents the number of samples per cycle, $F_s$ is the sampling frequency, and $F_{n}$ is the nominal frequency of the power system, which is assumed to be 60~{Hz}.

Generally, PQ event records capture several cycles of the voltage or current signal that occur before a disturbance begins. The DFRs that recorded the data used in this work are normally set to record fifteen cycles of data before a disturbance. The nominal peak values of voltage and current signals are determined using these 
``pre-event'' cycles for each processed signal. For this work, the first cycle in the event record is used to determine these nominal scalar values denoted as: (i) $\hat{V}_q$ for nominal peak voltage, (ii) $\hat{I}_q$ for nominal peak current, (iii) $\bar{V}_q$ for nominal Root Mean Square (RMS) voltage, and (iv) $\bar{I}_q$ for nominal RMS current. The magnitudes of voltage and current signals are compared to these nominal values to normalize the data with respect to the particular voltage or current level of the power system. This allows for more flexibility for these tools to be used at a different scale on the system.

\subsubsection{Root Mean Square}\label{sec:rms_calc}

The RMS of a signal is another characteristic used in the classification of electrical disturbance events. 
A signal's RMS is given by,
\begin{equation}
    \bar{x} = \sqrt{\dfrac{1}{N_{w}}\sum\limits_{i=1}^{N_{w}} \left|x[i]\right|^{2}},
    \label{eq:RMS}
\end{equation}
where $x$ is the analog signal, $N_{w}$ is the size of the RMS window, and $\bar{x}$ is the RMS calculation of the analog signal~\cite{bib:rms}. Unless otherwise stated, the size of the RMS window was set at the nominal number of samples in each cycle, $N_{c}$.

One use of RMS is in determining if the signal value is non-zero. In the instantaneous case, the sinusoidal signal will cross zero every half-cycle, so it is more difficult to tell whether the value remains near zero. A signal's RMS is used in events such as motor starting where the current increases over time.

\begin{figure}[!t]
\centerline{\includegraphics[width=\columnwidth]{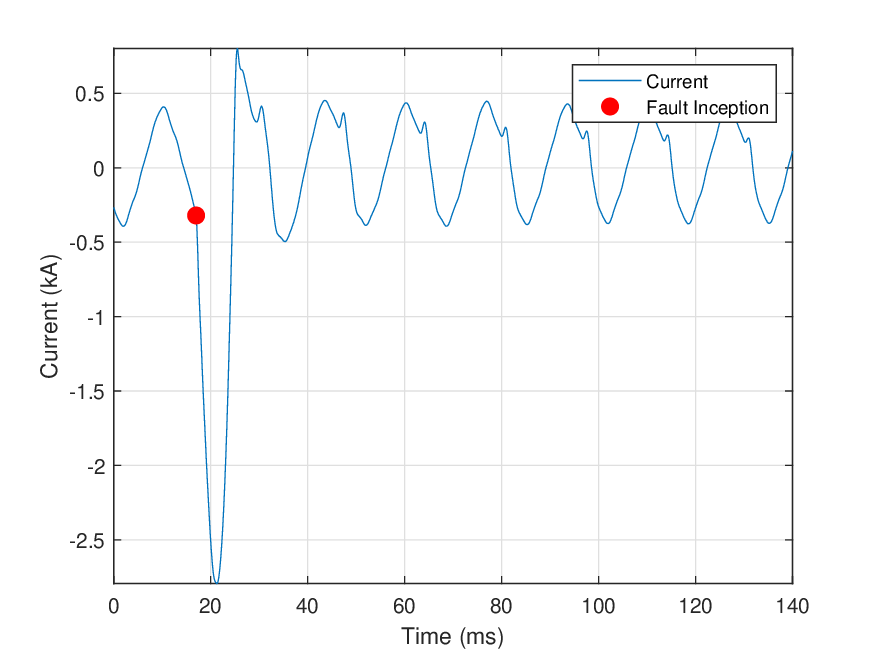}}
\caption{Fuse fault showing second derivative test}
\vspace{-5mm}
\label{fig:second_deriv}
\end{figure}

\subsubsection{Differentiation} \label{sec:differentiation}

One of the most common calculations used 
is a signal's derivative. 
The equation in \eqref{eq:deriv_sign_change} represents the first derivative with respect to the number of samples.

A positive first derivative indicates that the signal is increasing, and a negative first derivative indicates the signal is decreasing. This fact is used to detect the presence of 
peaks or spikes within a signal. The maximum or minimum of a peak or spike corresponds to the first derivative changing sign (i.e., going from positive to negative or vice versa). 
A change in the first derivative's sign is calculated by, 
\begin{equation}
    x'(n_1) \times x'(n_2)<0
\label{eq:deriv_sign_change}
\end{equation}
where $x'$ is the first derivative of the analog signal, $n_1$ is the sample before the first derivative's sign changes, 
and $n_2$ is the sample after the sign changes. Multiple sign changes 
over a short time interval provide a strong indication that a transient disturbance is present within the signal being processed. 

The second derivative is used to determine the change in the slope of the curve. A sudden increase in the second derivative shows as a sudden increase in slope and can indicate the point at which a fault begins. Fig.~\ref{fig:second_deriv} provides a representative illustration showing the use of the second derivative in determining the start of a fuse fault. The red circle shown is where the second derivative is higher than an empirical threshold, thus indicating a sudden increase in the slope of the curve. The third derivative is used to detect 
a shift in the slope of a curve.

\subsubsection{Harmonic Ratios} \label{sec:harmonics}

Harmonics can be key indicators of particular events within a transmission system (e.g., current transformer saturation, harmonic resonance, etc.). Harmonic analysis is facilitated through the calculation of the harmonic ratio, which is useful in determining the dominant frequency components within a signal. The $n^{\text{th}}$ harmonic ratio is calculated by,
%
\begin{equation}
    H_n =\dfrac{|X_n|}{|X_1|},
    \label{eq:harmonic}
\end{equation}
where $X$ is the Fast Fourier Transform (FFT) of $x$, $|X_1|$ is the magnitude of the fundamental frequency (i.e., 60~{Hz}), and $|X_n|$ is the magnitude of the $n^{\text{th}}$ multiple of the fundamental frequency~\cite{bib:wilson}.

\subsubsection{First Cycle Comparison} \label{sec:first_cycle}

\begin{figure}[!b]
\vspace{-5mm}
\centerline{\includegraphics[width=\columnwidth]{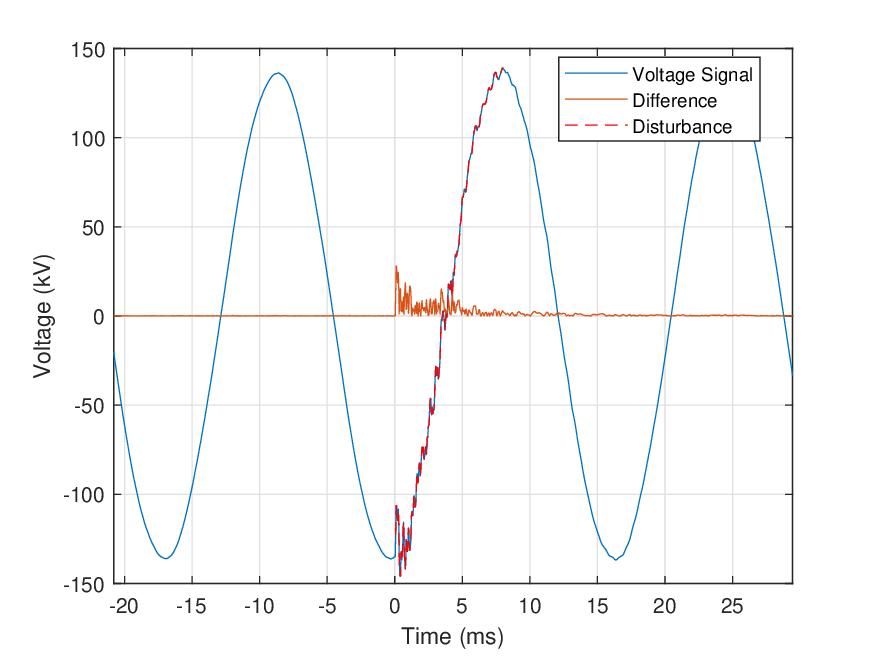}}
\caption{Voltage signal showing disturbance during capacitor switching.}
\label{fig:disturbance_finder}
\end{figure}

The CSV files generally store at least fifteen cycles of the voltage and current signals that occur prior to the disturbance event, thus 
a useful disturbance detection approach is to compare the signal's first cycle with each of its remaining cycles within the CSV file. 
After the first cycle is selected, it is replicated to construct an ideal signal that is of the same length as that of the recorded signal from which the first cycle was extracted. The generated ideal signal is then subtracted from the recorded signal. 
The time indices where this difference is very high indicates the start of a disturbance. Fig.~\ref{fig:disturbance_finder} illustrates the application of this approach in detecting the start of a 
capacitor switching event within a recorded voltage signal. Fig.~\ref{fig:disturbance_finder} shows the voltage signal with the capacitor switching disturbance portion of the signal highlighted and the result of the difference calculation overlaid. Where the difference calculation is highest corresponds with the start of the capacitor switching event, which is assigned a start time of zero milliseconds. 

\subsection{Continuous Signal Processing}

In addition to disturbance event classification this work makes use of the cyclic histogram in an attempt to reduce the memory storage requirements associated with 
a DFR's continuously recorded signal data. This work extends the cyclic histogram by also generating the residual and frequency histograms. 
The cyclic histogram was first proposed in~\cite{bib:cooke} to significantly reduce the size of continuously recorded oscillography data. This reduction in size allows for data to be stored for much longer than the OSG file, and allows for PQ analysts to pull data from each DFR without putting strain on the telecommunications network. A Python\textsuperscript{\textregistered} script 
performs the following tasks:

\begin{itemize}[leftmargin=*]
    \item Read the most recent configuration (CFG) file and extract the necessary data to read and correctly interpret the matching oscillography (OSG) file.
    \item Time-synchronize each cycle reliably to generate the cyclic and residual histograms.
    \item A custom ``maximum frequency'' calculation is developed to generate the frequency histogram that is faster and less computationally intense than traditional FFT processing.
    \item Generation of six CSV files. For each of the three histogram types, there is a CSV that contains the histogram and an accompanying metadata file that stores the bin values and record dates.
\end{itemize}

Signals are analyzed based on a sine representation, thus the continuous signal data is 
processed using a 
negative-to-positive transition in the cycle. This negative-to-positive transition is designated 
the beginning and end of each cycle. This helps in cases of signal disturbance as the disturbance is typically a magnitude disturbance and not additive. 
Current signal cyclic histograms are not generated due 
to transformers' inductive nature, 
which makes current an effect and more prone to its sinusoidal activity being 
negatively impacted 
to the point where cyclic analysis is not possible. 
Voltage is source driven, thus making it less susceptible to drift. 

The most recent CFG file is loaded and the OSG metadata extracted. The OSG metadata provides the: 
number of channels, sampling rate, and timestamp in accordance with IEEE COMmon format for TRAnsient Data Exchange (COMTRADE) Standard 2013~\cite{bib:comtrade}.

\subsubsection{Time Synchronization}
Due to the physical properties of the transmitted voltage, the signal is never exactly 60~{Hz} and the sampling Data Acquisition (DAQ) device will never sample the signal at the exact point of $x(t)=0$. At the transformer, the frequency can drift by as much as $\pm 0.03$~{Hz}, so the exact time in between cycles is not consistent. Due to this inconsistency, the position of $x(t)=0$ must be estimated to synchronize each cycle before generating the cyclic histogram. If this frequency drift is not account for, then it is impossible to generate the cyclic histogram for one hour of continuous oscillography data. 
Each cycle is detected
by finding two consecutive negative-to-positive transitions in the sampled waveform $x[n]$. A window is collected starting with the sample before the first transition, and the sample directly after the second transition and then processed for time synchronization. An ideal time vector $t_{I}$ is created as a reference where $t\in[0,1/F_{n}]$ in steps of $\Delta t$. A relative time $t_{r}$ vector is generated based on the slope estimated from the 
windowed signal. The first slope is,
\begin{equation}
    m_{1} = \dfrac{x[2]-x[1]}{\Delta t},~\text{ and } b_{1} = x[2] - m_{1}t[2],
\end{equation}
where $m_{1}$ is the slope between the first two sampled points and $b_{1}$ is the estimated position of the first zero-crossing. The first entry of the relative time vector is,
\begin{equation}
    t_{r}[1] = t_{I}[1]+\frac{b_{1}}{m_{1}}.
\end{equation}
The end of the windowed signal is used to find the second slope characteristics,
\begin{equation}
    m_{1} = \dfrac{x[N_{c}+1] - x[N_{c}]}{\Delta t}, b_{2} = x[N_{c}] - m_{2}*t_{I}[N_{c}].
\end{equation}
The last entry of the relative time vector is,
\begin{equation}
t_{r}[N_{c}+1] = t_{I}[N_{c}] - \left( \dfrac{-b_{2}}{m_{2}}-\dfrac{1}{F_{n}} \right).
\end{equation}
The rest of the relative time vector is,
\begin{equation}
    \Delta t_{r} = \dfrac{t_{r}[N_{c}+1]-t_{r}[1]}{N_{c}+1}.
\end{equation}
Now that the relative time vector has been calculated, the values of $x(t)=0$ lines up with $t_{r}=[0,1/F_{n}]$. Linear interpolation is used to generate a representation of the sampled waveform $x[n]$ from the relative $t_{r}$ and synchronize it onto the ideal time vector $t_{I}$. Once a cycle has been collected and synchronized, it is then stored to generate the cyclic and residual histograms.

\subsubsection{Histogram Generation}
The cyclic histogram is a combination of per-sample histograms concatenated to show the quality of the signal over time. For the case of $N_{c} = 16$, sixteen histograms are generated for each sample in the nominal cycle and stored in a matrix that represents the cyclic histogram. The global minimum and maximum of all of the synchronized cycles are used as the bin limits of all histograms to maintain a consistent scale for the cyclic histogram. Each histogram is generated using the $n^{\text{th}}$ sample of each of the synchronized cycles. 
By default, there are 1,024 bins per histogram, but this resolution can be increased or decreased as desired by utility personnel. 
A large number of bins will increase the size of the generated, output file. The cyclic histogram is generally unexciting as seen in Fig.~\ref{fig:cyclic_hist}. A residual histogram is generated by subtracting the first cycle from the remaining cycles in the record. Subtracting the first cycle accentuates any abnormal behavior(s) present within the processed signal at a per cycle resolution. 
The residual histogram--corresponding to the cyclic histogram in Fig.~\ref{fig:cyclic_hist}--is presented in Fig.~\ref{fig:cyclic_hist}. The voltage in Fig.~\ref{fig:residual_hist} is within the range of approximately $\pm135$~{kV} while the voltage range is $\pm4$~{kV} in the residual histogram of Fig.~\ref{fig:residual_hist}. This is almost a 40-times increase in activity resolution for no additional data cost.

\subsubsection{Frequency Histograms}
The dominant frequency is calculated using the FFT. 
Typically, the FFT is calculated over all frequencies within the range of 
$\pm{F_{s}/2}$. 
Calculating the FFT over this entire range of frequencies is inefficient, because the power grid's frequency is very stable with an expected maximum deviation of $\pm 0.03$~{Hz} with respect to the 60~{Hz} fundamental frequency. 
Based upon a sampling frequency of 960~{Hz}, a high resolution (i.e., a small step size between consecutive frequency values) frequency representation requires a significant number of zeros (e.g., 1.2 million) to be appended to the end of the time signal. Since the power grid's frequency is so stable most of the actionable information is contained within a very small range of frequencies, thus most of the resulting frequency response can be ``thrown out'' without loss of information. Zero padding the time signal--only to remove most of the resulting frequency response--represents a waste of computational resources and time. This problem is addressed by generating a support vector of frequencies centered at 60~{Hz} and with a Proccess BandWidth (PBW) of 0.2~{Hz}. 
The PBW can be changed based upon the specifics of: the DFR or equivalent device as well as utility personnel preferences or standards. 
The DFT of 
sixty cycles is calculated for only the frequencies specified in the support vector and 
a step size of thirty cycles between consecutive calculations. 
This results in the dominant frequency being calculated per second with an overlap of half a second. The DFT is calculated by,
%
\begin{equation}
    X[f] = \sum\limits_{n=1}^{N_{x}}{x[n]\exp\left[-j2\pi ft[n]\right]},
	\label{eqn:fft_calc}
\end{equation}
where 
\begin{equation*}
    f\in \left[F_{n}\pm \dfrac{PBW}{2}\right],
\end{equation*}
and $N_{x}$ is the total number of samples in the waveform over which the DFT is calculated~\cite{bib:oppenheim}.
The dominant frequency is selected by,
\begin{equation}
    F_{d}(t) = \operatorname*{arg\,max}_{f} |X[f]|.
\end{equation}
The output of the dominant frequency calculated for a sliding 60-cycle window of the recorded waveform and is stored and used to generate the frequency histogram. The support of the histogram is the same vector as the PBW calculated over in the DFT. The number of cycles per evaluation can be adjusted in the head of the code.

This process is accelerated using Python's \verb+mumba+ library that allows Just-In-Time (JIT) run-time compilation directly into machine code. Currently, JIT does not support the FFT algorithm, however it does support the calculation of the described, custom DFT. 
The result is not only faster, but requires far fewer computational resources and time than the zero padded FFT.

\subsection{Event Types}

\subsubsection{Current Transformer Saturation}

\begin{figure}[!t]
\centerline{\includegraphics[width=\columnwidth]{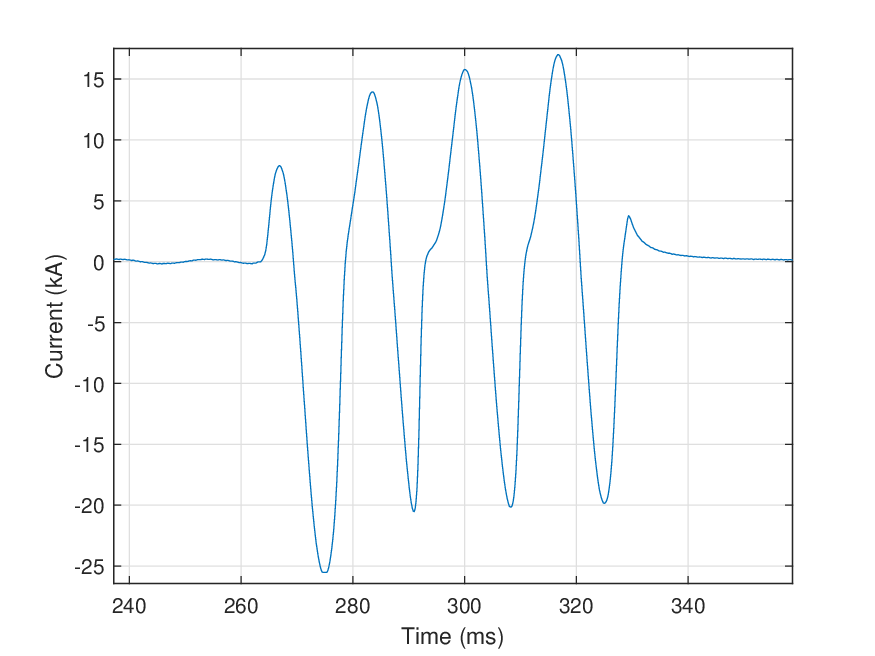}}
\caption{A representative illustration of ``kneeing'' within a current signal during a CT saturation event.}
\vspace{-5mm}
\label{fig:ct_sat}
\end{figure}

The first PQ event analyzed is Current Transformer (CT) saturation. A CT is commonly used in relaying or metering applications in high-voltage circuits by producing an alternating current in its secondary winding that is proportional to the current that it is measuring on the high-voltage system. These low-voltage, low magnitude currents are then used as input signals to various instrumentation~\cite{bib:ct_sat}. CT saturation occurs when the primary current is so high that its core 
cannot handle anymore flux. This results in inaccurate replication of the current signal on the secondary winding, which can cause protection relays to operate improperly. 
A key indicator of CT saturation is a change of slope as the current crosses zero each half-cycle. This change in slope is commonly referred to as ``kneeing''. Fig.~\ref{fig:ct_sat} shows a representative illustration of ``kneeing''--between 280~{ms} and 320~{ms}--within a CT's current signal.

In this work, the following criteria are used to determine the occurrence of CT saturation. These criteria are: (i) current exceeding fifteen times the continuous current rating of the CT, (ii) presence of DC offset, (iii) the DC offset returning to normal (i.e., 0~{Hz}) during the fault, (iii) inconsistent spacing between zero crossings, (iv) high third derivative of the current, (v) high second harmonic current, and (vi) high third harmonic within the current. A mix of these criteria are used to determine the likelihood of CT saturation as described at the end of this section.

The first step 
is to determine the presence of a fault or not. Processing continues if a fault is detected and moves to the next event otherwise. For the purposes of this work, a fault means that an abnormal flow of current has occurred causing the protective relay(s) to operate and trip the breaker(s). The presence of a fault is determined using the CT ratio defined in the COMTRADE configuration file. The CT ratio is,
\begin{equation}
    R_{\text{CT}}=\frac{I_P}{I_S},
    \label{eq:CTR}    
\end{equation}
where $R_{\text{CT}}$ is the turns ratio of the CT, $I_P$ is the rated continuous primary current, and $I_S$ is the rated continuous secondary current. The CTs in this work used a continuous rated current of 5~{Amperes} (A) on the secondary side of the CT. For instance, if the CT ratio $R_{\text{CT}}=240$, then the rated continuous current would be 1,200 A on the primary side and 5 A on the secondary side.

If the current exceeds fifteen times the continuous current rating of the CT, then a detected fault is high enough to be CT saturation. 
This threshold was empirically selected based upon recommendations of PQ engineers to ensure only abnormally high faults are selected since extremely high currents are generally indicative of CT saturation. Faults that do not meet this threshold will have a lower chance of being CT saturation. The threshold is given by,
\begin{equation}
    \dfrac{I(n)}{I_P} > \tau_{\text{CT}}, \label{eq:service_factor}    
\end{equation}
where $I$ is the instantaneous current being analyzed, $I_P$ is the rating of the CT on the primary side, $\tau_{\text{CT}} = 15$ is the CT saturation threshold, and $n = 1, 2, \dots, N$. The CT saturation threshold was set based upon inputs from power utility personnel, but can be changed based upon local criteria.

The presence of DC offset is also an indicator of CT saturation~\cite{bib:ct_sat}. For this particular event, DC offset is determined by first calculating the peak value of each cycle of the faulted section of the waveform. The peaks of the positive and negative half-cycles are then averaged together to give a value for the offset above or below 0~{A}. If the maximum of this value exceeds a threshold compared to the nominal peak current, then DC offset is detected in the fault as given by,

\begin{equation}
    \dfrac{|I_{\text{DC}}|}{\hat{I}_q} > \tau_{\text{DC}}, \label{eq:ct_dc_offset}    
\end{equation}
where $I_{\text{DC}}$ is the maximum DC offset detected during the fault, $\hat{I}_q$ is the nominal peak current extracted from the first cycle, and $\tau_{\text{DC}} = 3$ is the empirically selected threshold for the ratio of DC offset magnitude to nominal peak current. A loss of DC offset is detected if the offset magnitude is lower at the end of the fault than the beginning.

The number of samples between zero crossings is then compared to half the nominal number of samples in each cycle calculated using \eqref{eq:spc} as described in Sect.~\ref{sec:nominal_values}. The zero crossing points are calculated as the indices at which the waveform changes sign (i.e., from negative to positive or vice versa). The number of samples between each zero crossing is calculated for every cycle by subtracting the indices accordingly. This number of samples is compared to the nominal value and is given by,
\begin{equation}
   \max\left|N_{\text{Z}}(k) - \dfrac{N_c}{2}\right| > \tau_{\text{Z}},\;k=(1,2,3, \dots, N_{\text{F}})
    \label{eq:zero_cross}    
\end{equation}
where $N_\text{Z}$ is the number of samples between zero crossings, $N_c$ is the nominal number of samples in each cycle, $k$ is index of each cycle, $N_{\text{F}}$ is the total number of cycles in the faulted portion of the waveform, and $\tau_{\text{Z}}=10$ is the empirically selected threshold for the difference from nominal in the number of zero crossings.

The ``kneeing'' present in the waveform is detected using a third derivative test. The maximum third derivative present in the first cycle of the waveform (i.e., before the fault) is used as the nominal value. The maximum third derivative of the faulted portion of the waveform is compared to the nominal value and will be ``flagged'' if it exceeds a certain threshold as given by,

\begin{equation}
    \dfrac{\max|{I_f'''(n)}|}{\max|{I_{c}'''(n)}|}>\tau_{\text{D3}}
    \label{eq:kneeing}
\end{equation}
where $I_f'''(n)$ is the third derivative of the faulted current waveform, $I_c'''(n)$ is the third derivative of the first cycle of the current signal, and $\tau_{\text{D3}}=5$ is the empirically selected threshold for the ratio of the fault third derivative with the nominal one.

Finally, the 
the harmonic ratios of the entire current waveform are calculated using equation~\eqref{eq:harmonic} as described in Sect.~\ref{sec:harmonics}. A very good indicator of CT saturation is when the 
second and third harmonic currents exceed 
the thresholds of \textbf{15\%} and \textbf{5\%} of the fundamental, respectively. 

All these criteria are combined to give a confidence level for CT saturation as given by:
\begin{itemize}[leftmargin=*]
    \item \textit{High confidence:} The thresholds are exceeded for the current rating of the CT and the second harmonic current. The thresholds must also be exceeded for \textit{three} of the following: DC offset, loss of DC offset, inconsistent spacing between zero crossings, third derivative, or third harmonic current.
    \item \textit{Medium confidence:} The threshold is exceeded for the current rating of the CT, but the second harmonic threshold is not exceeded. The thresholds must then be exceeded for \textit{three} of the following: DC offset, loss of DC offset, inconsistent spacing between zero crossings, third derivative, or third harmonic current.
    \item \textit{Low confidence:} The threshold is exceeded for the current rating of the CT, but the second harmonic threshold is not exceeded. The thresholds must then be exceeded for \textit{two} of the following: DC offset, loss of DC offset, inconsistent spacing between zero crossings, third derivative, or third harmonic current.
    \item \textit{Low confidence (alternative):} The threshold is not exceeded for the current rating of the CT but is for the second and third harmonics. The thresholds must then be exceeded for \textit{two} of the following: DC offset, loss of DC offset, inconsistent spacing between zero crossings, or third derivative.
\end{itemize}

\subsubsection{Analog-to-Digital Converter Clipping}

\begin{figure}
\centerline{\includegraphics[width=\columnwidth]{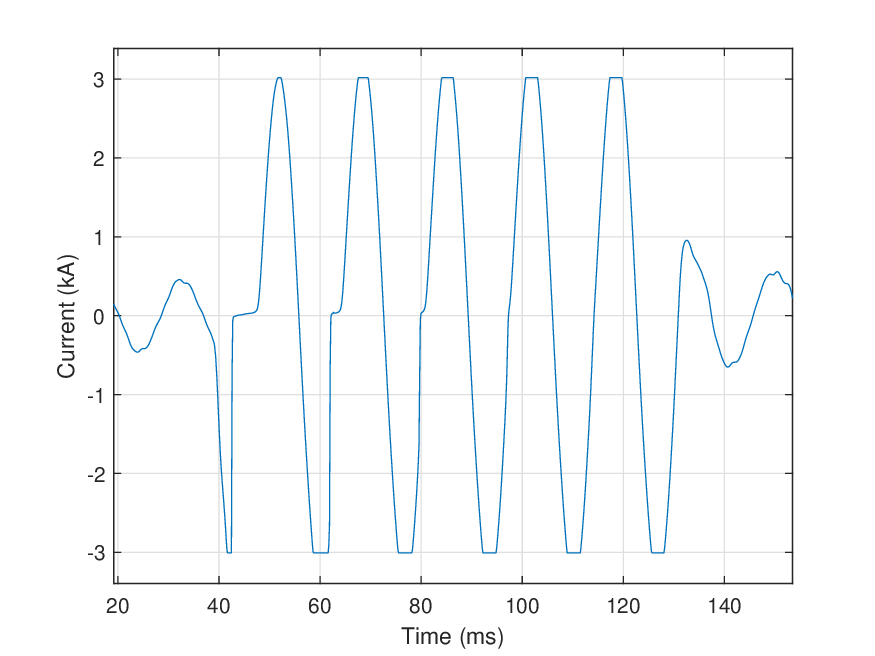}}
\caption{A representative current signal showing Analog-to-Digital Converter (A/D) clipping.}
\vspace{-5mm}
\label{fig:ad_clipping}
\end{figure}

An analog-to-digital (A/D) converter is a device that converts continuously varying analog signals into a binary or digitized sequence. Many electronic devices in substations (e.g.,relays and DFRs) utilize A/D converters to record voltage and current signals in a binary format. The range of the digitized scale is restricted by the power supply rail voltage. If the analog value results in a digitized sequence that exceeds the rail voltage, then the digitized sequence will appear ``clipped'' or ``flat-topped'' at its minimum and maximum values. For substation devices, clipping often appears in 
current signals during fault events. This results in inaccurate replication of the current signals, which can result in relaying mis-operation. Fig.~\ref{fig:ad_clipping} shows the visible clipping at the minimum and maximum values of a current signal's digitized sequence.

Clipping is indicated by the repetition of equal magnitude samples within the digitized sequence. 
First, the index of the absolute maximum of the signal is calculated. The section of the waveform \textbf{ten} samples before and \textbf{ten} samples after the maximum is then extracted for analysis. If the first derivative of this section of the signal is equal to zero for more than \textbf{four} consecutive samples, then A/D converter clipping is present within the signal. 

\subsubsection{Induced Transient Noise due to Switching} \label{sec:noise}

\begin{figure}[b]\vspace{-5mm}
\centerline{\includegraphics[width=\columnwidth]{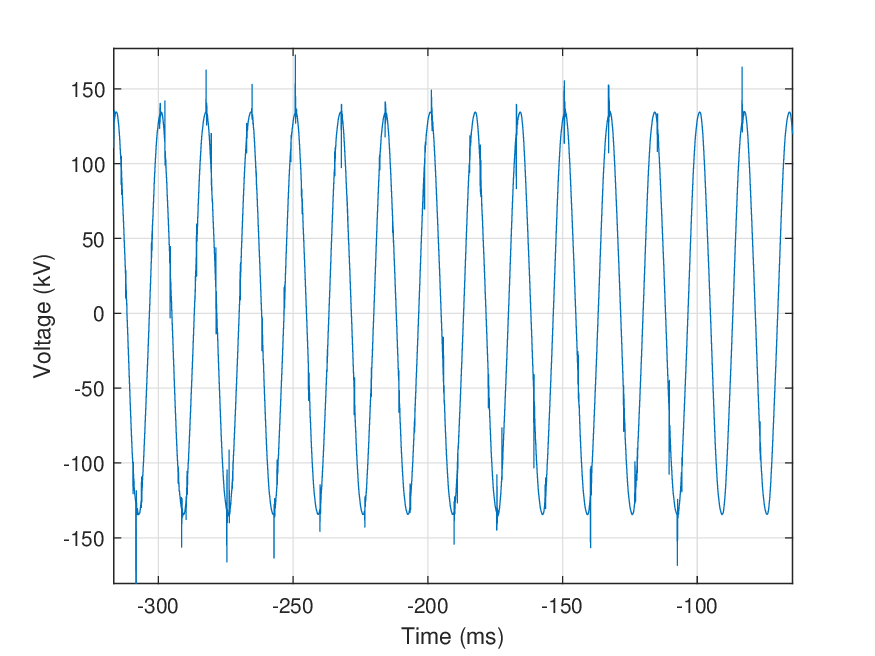}}
\caption{ A representative voltage signal showing transient noise due to switching.}
\label{fig:transient_noise}
\end{figure}

When high voltage devices--such as air-break switches--are opened to de-energize a bus section, 
the resulting arcing can induce 
high-frequency noise 
upon the voltage or current signals of the electronic monitoring equipment (e.g., PQ monitor). Identification of this induced 
transient noise is used to determine where 
signal chokes may need to be installed or where shielding and ground bonding integrity may need to be checked. 
Fig.~\ref{fig:transient_noise} provides a representative illustration of this transient noise within a voltage signal.

This 
event is characterized by the presence of small random spikes (i.e., noise) 
throughout the voltage or current signals. Switching induced transient noise is identified by its: (i) overall difference from an ideal waveform, (ii) harmonic content below \textbf{5\%} of the fundamental, (iii) sudden spikes determined by the first derivative exceeding \textbf{10\%} of the nominal peak value, (iv) persistence over \textbf{five} cycles or more, (v) occurrence averaging \textbf{once} per cycle, (vi) instances totaling \textbf{twenty} or more, and (vii) presence causing individual sample values to exceed the nominal peak signal value occurring at least \textbf{five} times.

The first criterion is determined using the approach described in Sect.~\ref{sec:first_cycle} in which a voltage signal is compared to a reference signal, which is made up of replications of the first cycle. The condition in which the difference between the actual voltage and the reference voltage exceeds a threshold is given by,
\begin{equation}
    \dfrac{\bar{V}_{\Delta}}{N}>\tau_{\text{N}}
    \label{eq:noise_diff}
\end{equation}
where $\bar{V}_{\Delta}$ is the mean value of the voltage difference between the actual and ideal signals, $N$ is the total number of samples in the waveform, and $\tau_{\text{N}}=30$ is the empirically chosen threshold 
for this ratio. If the first six criteria are met, then induced transient switching is classified with \textit{medium} confidence. If all seven criteria are met, then this event type is classified with \textit{high} confidence.

\subsubsection{High-Speed Reclosing with Tapped Motor Loads}

\begin{figure}[!t]
\centerline{\includegraphics[width=\columnwidth]{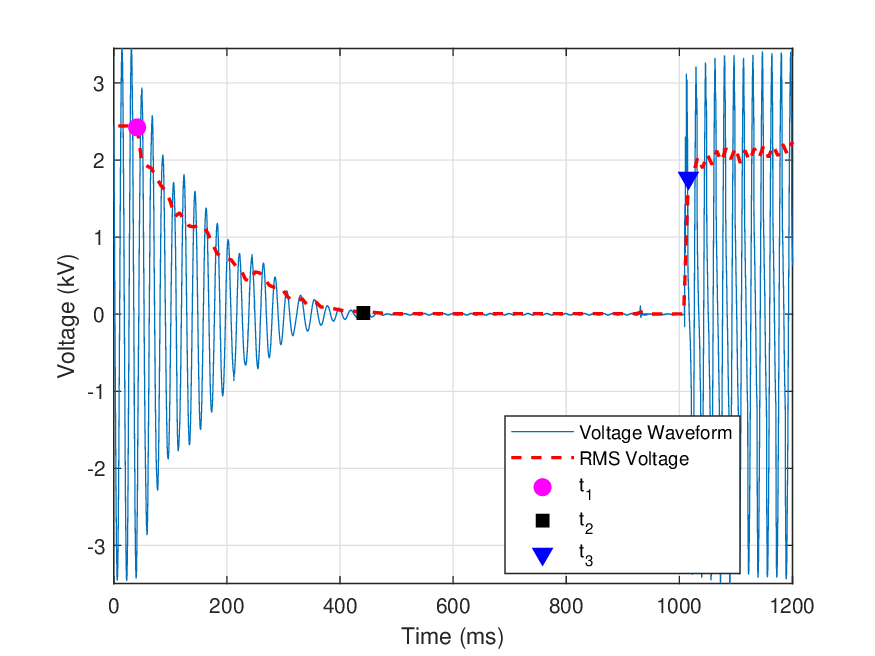}}
\caption{A representation of the case in which the voltage signal \textit{does} decay sufficiently prior to a successful reclosing operation in the presence of a tapped motor load.}
\vspace{-5mm}
\label{fig:hs_reclosing1}
\end{figure}

A common practice is to employ high-speed instantaneous reclosing on faulted transmission lines. Sometimes there may be large or significant motor load served from stations tapped on the line. For this work, a motor load is considered significant if it is directly served from a high-voltage transmission line (e.g., 161~{kV}). In such cases, the line voltage may be supported by the motors--as they spin down--so that residual voltage remains on the line by the time of a high-speed breaker reclose operation. The residual voltage may require up to five seconds to decay in large machines~\cite{bib:patterson}. Since this residual voltage is unlikely to be in phase with the system voltage, the result can be a failed reclose attempt by 
the line breakers as well as damage to the motors. 
Thus, it is important to identify lines where high-speed reclosing needs to be delayed to allow the voltage to sufficiently decay before carrying out the reclosing operation. Fig.~\ref{fig:hs_reclosing1} shows a voltage signal in which sufficient time has passed to allow the voltage signal to decay to a point after which the reclosing operation was successfully completed. 

For identification of this event, it must be determined whether the reclosing operation is a high-speed reclosing operation. 
For this work, the reclosing operation is a high-speed one if it is ``blind'' (i.e., without any supervision or checks) and occurs within thirty cycles of the initial current interruption by the breaker~\cite{bib:patterson}. Identification of the reclosing with tapped motor loads is achieved by determining the sample points at which the: (i) voltage signal begins to decay, (ii) voltage signal reaches zero, and (iii) reclosing operation occurred. The time between these three points determines whether the reclosing is a high-speed operation. In this work and as shown in Fig.~\ref{fig:hs_reclosing1}, these three sample points are designated as $t_{1}$ (magenta circle), $t_{2}$ (black square), and $t_{3}$ (blue triangle), respectively. The location of these three sample points is determined using the RMS signal, which is calculated using equation \eqref{eq:RMS} as described in Sect.~\ref{sec:rms_calc} and is shown in Fig.~\ref{fig:hs_reclosing1} as a broken, red line. For this event, the RMS window was set to half the number of samples in each cycle (i.e., $N_c/2$).

The point $t_{1}$ is the time at which the RMS voltage first decays below a threshold and is determined by, 
\begin{equation}
     \dfrac{\bar{V}(t)}{\bar{V}_q(t)} < \tau_{\text{S}},
    \label{eq:reclose_sag}
\end{equation}
where $\bar{V}$ is the RMS of the voltage, $\bar{V}_q$ is the nominal RMS voltage as determined from the first cycle, and $\tau_{\text{S}}=0.9$ is the empirically selected threshold for the sag in voltage indicating the start of a decay. The point $t_{2}$ is determined as the time at which the voltage decays low enough to be considered approximately zero. An empirical threshold of $\tau_{\text{0}}=0.01$ was used as the threshold below which the RMS voltage must reach to be considered zero. If this condition is not met, then $t_2$ is the time at which the RMS voltage is at its minimum. The RMS voltage must decay to below \textbf{50\%} of the nominal value for the process to continue.

The voltage decay portion is the RMS voltage between times $t_1$ and $t_2$ and is designated here as $\bar{V}_{\text{D}}$. The median (i.e., middle value) of $\bar{V}_{\text{D}}$ must be lower in magnitude than the voltage at time $t_1$ and higher than the voltage at time $t_2$. The mean of the first derivative of $\bar{V}_{\text{D}}$ must also be negative to indicate a downward slope or decrease in voltage. The maximum first derivative of the voltage decay must also be less than a threshold to ensure that the voltage decay was not sudden. This condition is given by,
\begin{equation}
    \dfrac{\max|\bar{V}_{\text{D}}'|}{\bar{V}_q}<\tau_{l}
    \label{eq:reclose_step_down}
\end{equation}
where $\bar{V}_{\text{D}}'$ is the first derivative of the decaying portion of the RMS voltage, $\bar{V}_q$ is the nominal RMS voltage, and $\tau_{l}=0.5$ is the empirically selected threshold for the maximum first derivative of the decaying voltage. The point $t_3$ is the time at which the RMS voltage increases by \textbf{30\%} of nominal value in one RMS sample. This condition is determined by the first derivative of the RMS signal as given by,
\begin{equation}
    \dfrac{\max|\bar{V}_{\text{S}}'|}{\bar{V}_q}>\tau_{\text{U}}
    \label{eq:reclose_step_up}
\end{equation}
where $\bar{V}_{\text{S}}'$ is the first derivative of the portion of the RMS voltage after time $t_2$, $\bar{V}_q$ is the nominal RMS voltage, and $\tau_{\text{S}}=0.3$ is the empirically selected threshold for the minimum first derivative of the reclosing voltage. Time $t_3$ is the point when reclosing occurs and the voltage is restored.

\begin{figure}[!t]
\centerline{\includegraphics[width=\columnwidth]{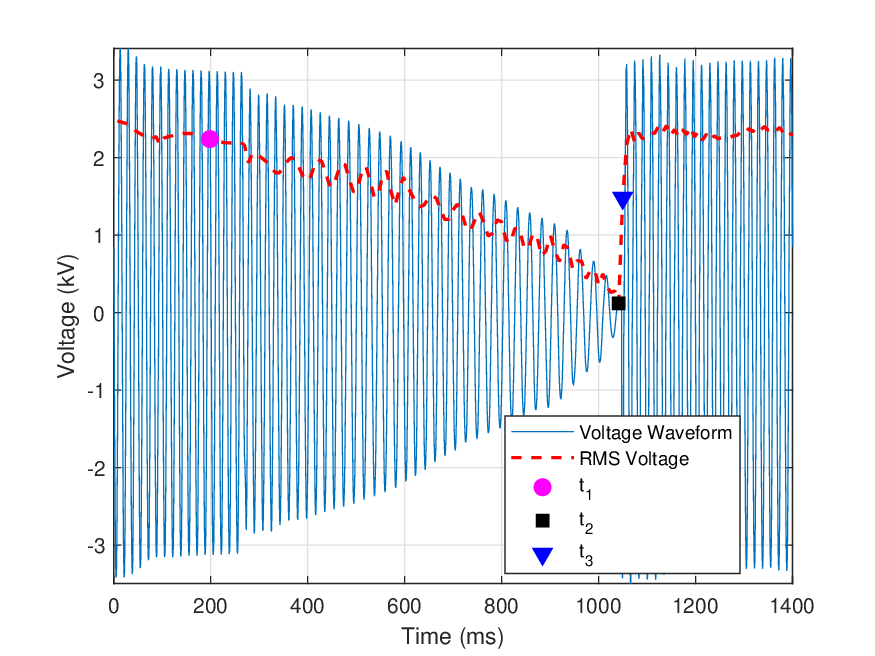}}
\caption{
A representation of the case in which the voltage signal \textit{does not} decay sufficiently prior to a successful reclosing operation in the presence of a 
tapped motor load.}
\vspace{-5mm}
\label{fig:hs_reclosing2}
\end{figure}

The criteria given thus far serve to classify the event as normal reclosing with a tapped motor load. Fig.~\ref{fig:hs_reclosing1} is a normal event in which there was sufficient time between $t_2$ and $t_3$. If there is not sufficient time between these two points, then the event is ``flagged'' as needing attention. The condition that defines a high-speed reclosing operation is given by,
\begin{equation}
    t_3-t_2>\tau_{\text{HS}}
    \label{eq:reclose_high_speed}
\end{equation}
where $t_2$ is the time at which the voltage first decays to zero, $t_3$ is the time at which the voltage is restored, and $\tau_{\text{HS}}=30~\text{cycles}$ is the threshold for the minimum time the voltage must be zero before reclosing as recommended~\cite{bib:patterson}. 
Fig.~\ref{fig:hs_reclosing2} shows a case in which the minimum time for which the voltages needs to be zero is not satisfied.

\subsubsection{DC Offset}

DC offsets in analog channels are a common issue and when they are large enough can negatively impact RMS calculations. A large DC offset is accounted for by re-calibration of the corresponding monitoring or recording device. Automated calculation of DC offset affords utility personnel the ability to prioritize re-calibration of those devices associated with the largest amounts of DC offset. 
The DC offset event is characterized by an asymmetry between the positive and negative half-cycles of a voltage or current signal.

\begin{figure}[!t]
\centerline{\includegraphics[width=\columnwidth]{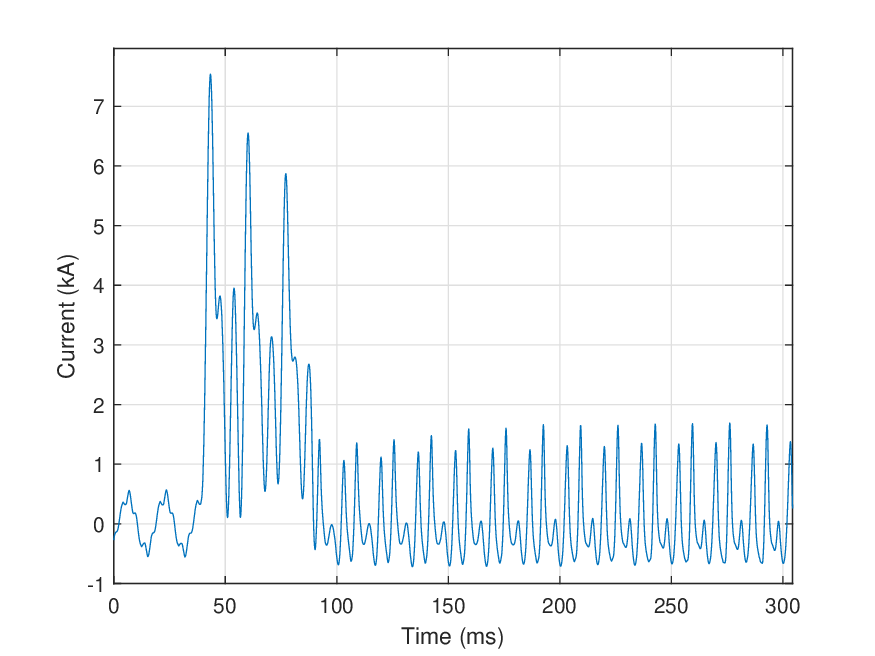}}
\caption{Representative illustration of a large DC offset--from 40 ms to 90 ms--within a current signal.}\vspace{-5mm}
\label{fig:dc_offset}
\end{figure}

The presence and amount of DC offset is determined using both time and frequency domain analysis. In the frequency domain, a DC offset is present if the magnitude of the 0~{Hz} frequency component is greater than 50\% of the magnitude at the fundamental frequency component (i.e., 60~{Hz} in the United States). Mathematically this condition can be expressed as,
\begin{equation}
    \dfrac{X_0}{X_1} > \tau_{f}
    \label{eq:dc_harmonics}
\end{equation}

\begin{figure}[!b]
\vspace{-5mm}
\centerline{\includegraphics[width=\columnwidth]{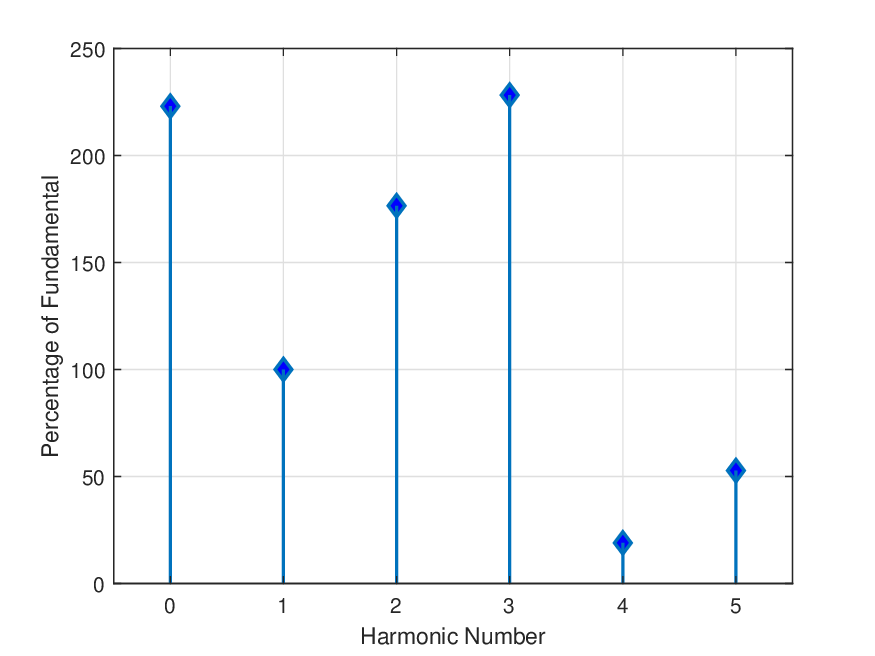}}
\caption{Illustration of the zeroth through fifth harmonic ratios of current signal shown in Fig.~\ref{fig:dc_offset}.}
\label{fig:dc_stem}
\end{figure}

\noindent where $X_0$ is the magnitude of the 0~{Hz} frequency component, $X_1$ is the magnitude at the fundamental frequency component, and $\tau_{f} = 0.5$ is empirically selected as the minimum ratio with respect to the fundamental frequency. Fig.~\ref{fig:dc_offset} provides a representative illustration of a current signal in which a large amount of DC offset is present from 40~{ms} to 90~{ms}. Fig.~\ref{fig:dc_stem} shows the magnitude of the zeroth through fifth harmonic of the current signal shown in Fig.~\ref{fig:dc_offset}. In this case, the 0~{Hz} frequency component is over two times larger than that of the fundamental frequency component (i.e., the first harmonic) and would be ``flagged'' as a DC offset event. Interestingly, the presence of the third harmonic indicates that another disturbance is also present within the recorded signal of Fig.~\ref{fig:dc_offset}.
%

If the frequency domain analysis results in the identification of a DC offset event, then time domain analysis is performed as a validation step. Time domain analysis is conducted by computing the mean over each cycle within the recorded signal. If a given cycle's mean value is zero, then there is no DC offset present within that cycle. This is because the area under the positive and negative portions of the cycle would negate each other. However, if the selected cycle's mean exceeds 50\% of the nominal signal's peak value, then the DC offset event ``flag'' is set once more. The amount of DC offset--returned by the automated process--is,
\begin{equation}
    \operatorname*{arg\,max}_{i} \mu_{i},
\end{equation}
where $\mu_{i}$ is the mean value of the $i^{\text{th}}$ cycle within the signal being processed. 

\subsubsection{Motor Starting}

\begin{figure}[!t]
\begin{subfigure}{\columnwidth}
    \centering
    \includegraphics[width=\columnwidth]{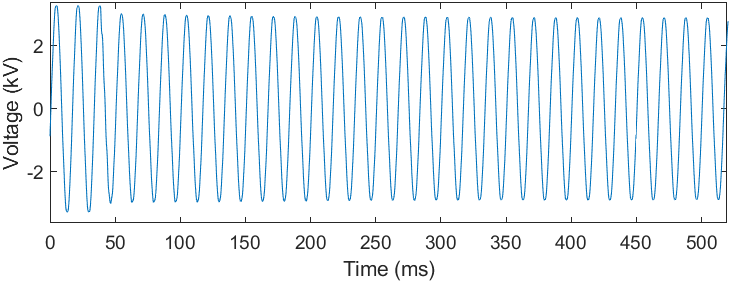}
    \caption{Voltage signal.}
    \label{fig:motor_voltage}
    \end{subfigure}
    
    \begin{subfigure}{\columnwidth}
    \centering
    \includegraphics[width=\columnwidth]{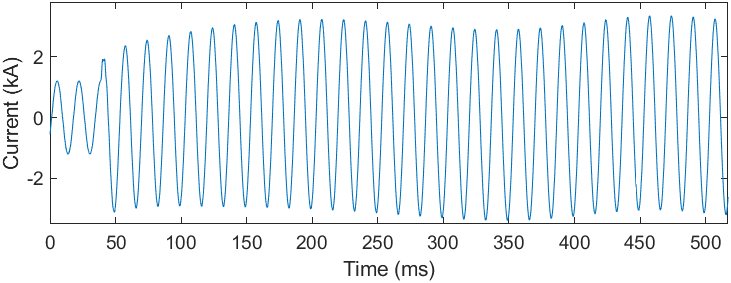}
    \caption{Current signal.}
    \label{fig:motor_current}
\end{subfigure}
\centering
\caption{Voltage and current signals showing signal characteristics associated with a motor starting event.}
\vspace{-5mm}
\label{fig:motor_start}
\end{figure}

Instantaneous increases in current may be due to faults, motor starts, transformer energizations, or other events. Signatures present within the recorded signals 
can be used to distinguish and classify each of these events. PQ disturbances can then be correlated by event classification. In the case of motor starting, the voltage sags 
and the current can increase to five to six times its rated value~\cite{bib:epri}. It is challenging 
to set protective relays in such a way to enable recognition of a motor starting event rather than recognizing the event as a fault on the system. 
The automated processed described in this section is developed under the assumption that the corresponding relays are properly set so they do not trip open when motor inrush current is present. Fig.~\ref{fig:motor_voltage} and Fig.~\ref{fig:motor_current} show representative illustrations of motor starting voltage and current signals, respectively. 

The automated process checks for a 
voltage sag below 95\% of the signal's nominal RMS value and a current spike to twice the CT's rated value determined by~\eqref{eq:CTR}. If both of these conditions persist for at least ten consecutive cycles, then the first indicator of motor starting is identified. 
The persistence of both conditions--for ten or more consecutive cycles--distinguishes motor starting events from a fault condition, which typically occurs for only several cycles before the relay trips open the breaker. Motor starting events are also associated with a frequency response that is low in harmonic content. Thus, if none of the voltage or current signals' harmonics exceed \textbf{15\%} of the fundamental frequency components magnitude, then the second indicator of motor starting is identified. The final indicator for motor starting is that all three conditions (i.e., voltage sag, current spike, and harmonics below 15\% of the fundamental) occur on all three phases, because motors are three phase devices.

\subsubsection{Variable Frequency Drive Motor Starting}

\begin{figure}[!t]
\centerline{\includegraphics[width=\columnwidth]{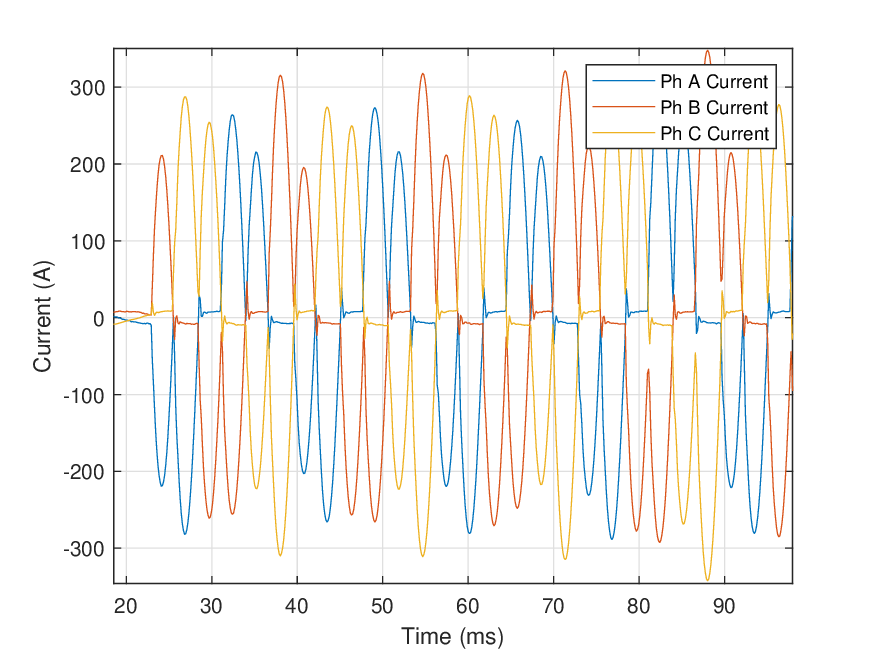}}
\caption{A illustration showing the distinct current signal generated during a six-pulse VFD motor start event.}
\vspace{-5mm}
\label{fig:vfd_start}
\end{figure}

Some motors utilize electronic starting (e.g., Variable Frequency Drives -- VFDs) to bring the motor up to speed in a controlled manner to limit voltage supply disturbance(s). 
VFDs produce unique harmonic patterns, which allows these events to be easily identified by our automated process. 
When a VFD motor starts it creates a very distinct current signal. A representative illustration of this distinct current signal can be seen in Fig.~\ref{fig:vfd_start}.

\begin{figure}[!t]
\centerline{\includegraphics[width=\columnwidth]{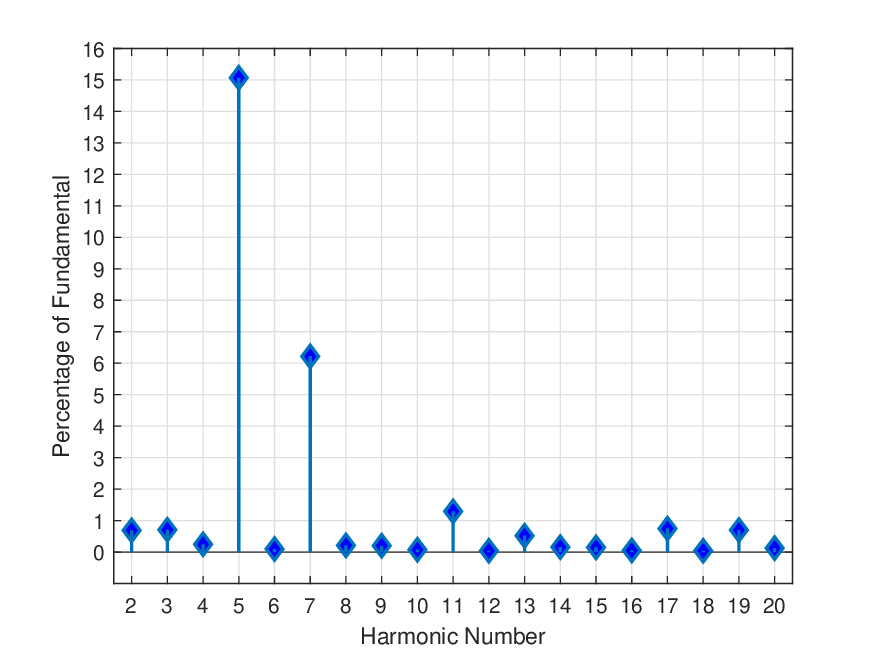}}
\caption{The harmonic ratios calculated from the current signal of the six-pulse VFD motor start event shown in Fig.~\ref{fig:vfd_start}.}
\vspace{-5mm}
\label{fig:vfd_harmonics}
\end{figure}

In Fig.~\ref{fig:vfd_start}, each phase has two pulses per half-cycle. The 
number of pulses per half-cycle indicates the type of VFD 
(e.g., six-pulse, twelve-pulse, etc.), thus VFD motor starting events are identified by counting the number of times the current signal drops below 50\% of each cycle's maximum value. Two pulses in each half cycle of a current signal for each phase (e.g., Fig.~\ref{fig:vfd_start}) would indicate a six-pulse VFD. The number of pulses for the drive is given by,
\begin{equation}
    N_p = \dfrac{3}{2}\times mode(K),\;K>2
    \label{eq:vfd_pulse}
\end{equation}
where $K$ is number of times the current crosses 50\% of each cycle's maximum value 
every half-cycle, and $mode(K)$ refers to the most often occurring value of $K$. 
The current must cross the threshold more than two times for at least \textbf{eight} cycles during the event to be considered VFD motor starting. After $N_p$ is calculated, harmonic analysis is conducted, because VFD motor starting events result in 
dominant harmonics 
on either side of an integer multiple of $N_p$. Fig.~\ref{fig:vfd_harmonics} shows the harmonics for the six-pulse (i.e., $N_p=6$) VFD motor starting event illustrated in Fig.~\ref{fig:vfd_start}. The fifth and seventh harmonics are the two most dominant harmonics and occur on either side of the sixth harmonic, which is equal to that of $N_{p} = 6$. 
The value of $N_{p}$ is validated by ensuring that the dominant harmonics are at least \textbf{five} times larger than the value of the harmonics at integer multiples of the $N_p$. This validation check is performed by,
\begin{equation}
    \frac{H_{kN_p\pm{1}}}{H_{kN_p}}>\tau_{\text{V}},\;(k=1,2,3,4)
    \label{eq:vfd_harms}
\end{equation}
where $N_p$ is the number of pulses in the VFD, $H_{kN_p}$ is the harmonic at an integer multiple of $N_p$, $k$ is an integer, and $\tau_{\text{V}}=5$ is the empirically determined threshold for the ratio of the dominant harmonics with those at integer multiples of $N_p$. If equation~\eqref{eq:vfd_harms} is satisfied, then 
the number of predicted pulses is deemed correct. Finally, the event is identified as VFD motor starting so long as all three currents (i.e., phase A, B, and C) increase over the events duration.

\subsubsection{Melting Fuse}\label{sec:fuse_melt}

\begin{figure}[b]\vspace{-5mm}
\centerline{\includegraphics[width=\columnwidth]{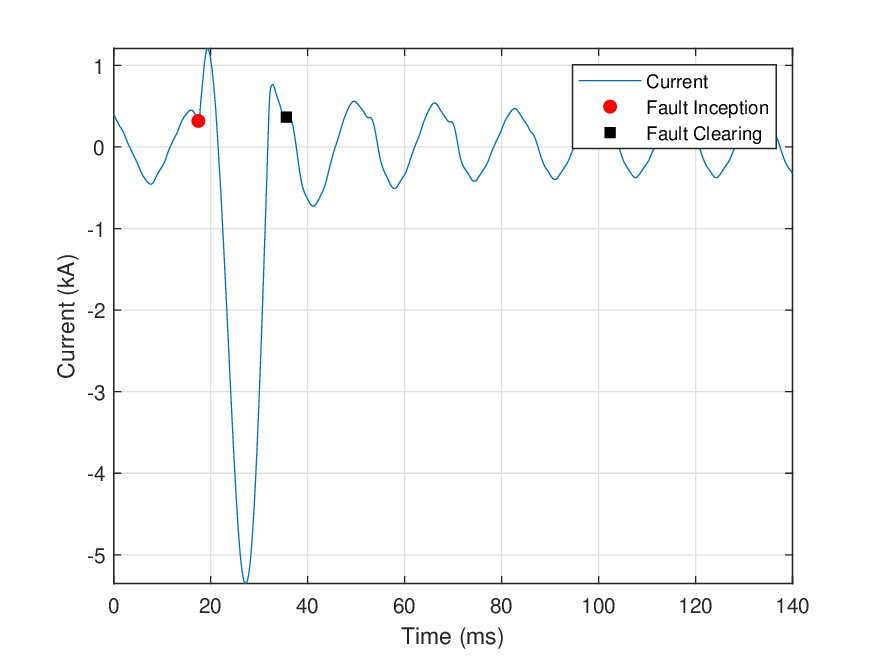}}
\caption{A current signal during a fuse melting event that last just over one cycle.}
\label{fig:fuse}
\end{figure}

Unlike a breaker, a blown (a.k.a., melted) fuse requires utility personnel to physically replace it, so it is helpful to distinguish fuse faults from breaker faults. 
These two faults are distinguished from one another by the speed at which the fault is cleared. Breakers require between two or more 
cycles to clear a fault while fuses require less than two cycles. 
Fig.~\ref{fig:fuse} shows an example of a fuse melting event that is cleared in a little more than one cycle. 

The key to automated identification of fuse melting events is accurate determination of the fault's 
inception and clearing points. A fuse melting event occurs if the total clearing time was less than one and a half-cycles and is determined by,
\begin{equation}
    |t_I - t_C| < \tau_c
    \label{eq:fuse_clearing}
\end{equation}
where $t_I$ is the inception point, $t_C$ is the clearing point, and $\tau_c = 1.5$ is the threshold for the maximum fuse clearing time.

Automated identification of a fuse melting event is initialized by determining if the event persisted for at least a quarter of a cycle and the current reaches at least twice its nominal value over the event's duration. The cycle before and just after the portion associated with these two conditions is then analyzed one half-cycle at a time to determine the fault inception and clearing points. The three possible approaches used to determine these points are: (i) a sign change in the first derivative, (ii) a sudden increase in the second derivative, and (iii) the current signal's zero crossings.

The first derivative approach is implemented using equation \eqref{eq:deriv_sign_change} as described in Sect.~\ref{sec:differentiation}. A sign change in the first derivative before or after the spike in current indicates the fault inception and clearing points. This approach is used to determine the inception and clearing points of the fuse melting event shown in Fig.~\ref{fig:fuse} where the red circle indicates the fault inception point, and the black square indicates the fault clearing point.

If the first derivative approach is unsuccessful (i.e., a sign change in the first derivative does not exist), then the second derivative is used as described in Sect.~\ref{sec:differentiation}. The condition for a large second derivative is given by,
\begin{equation}
    \dfrac{\max|{I}''(n)|}{\hat{I}_q}>\tau_{\text{D2}}
    \label{eq:fuse_second_deriv}
\end{equation}
where $I''(n)$ is the second derivative of the current signal, $\bar{x}_{c}$ is the nominal peak current, and $\tau_{\text{D2}}=0.02$ is the empirically selected threshold for the minimum ratio of the second derivative of the current to the nominal value. This approach was used to determine the fault inception point of Fig.~\ref{fig:second_deriv} as described in Sect.~\ref{sec:differentiation}. 

If the second derivative approach is also unsuccessful (i.e., the minimum threshold is not met), then the fault inception and clearing points are assumed to be the zero crossings just before and just after the current spike, respectively. After the fault inception and clearing points are determined, equation \eqref{eq:fuse_clearing} is used to determine whether the fault was short enough in duration to be a melted fuse.

\subsubsection{Ferroresonance}

\begin{figure}[!t]
\centerline{\includegraphics[width=\columnwidth]{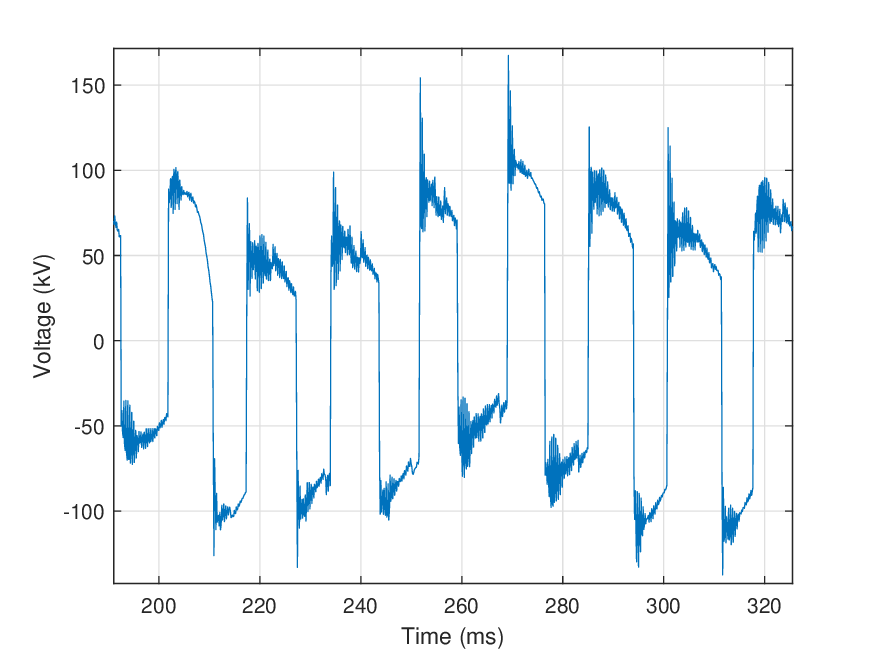}}
\caption{Illustration of a voltage signal collected during a  ferroresonance event.}
\vspace{-5mm}
\label{fig:ferroresonance}
\end{figure}

Ferroresonance is 
electric circuit resonance 
that occurs when a circuit containing a nonlinear inductance is fed from a source that has a series capacitance connected to it. In a transmission system, ferroresonance can occur when a breaker--with grading capacitors--is used to de-energize a bus that has magnetic Voltage Transformers (VTs) connected to it. 
The described scenario presents a serious safety risk to utility personnel and damage risk to equipment, because severe overvoltages can occur despite the breaker being in an open state. 
Ferroresonance manifests in the voltage signals and causes the signals to take on a square wave like shape/appearance. Fig.~\ref{fig:ferroresonance} provides a representative illustration of the square wave appearance that a voltage signal can take on due to ferroresonance. Another characteristic of ferroresonance events is that the current is normally zero during the event. This is due to the line being de-energized; however, depending on the recording device's location, the current can be recorded as a nominal signal.

Ferrroresonance events are identified using three criteria: (i) a large difference between discrete samples in the voltage signal, (ii) this behavior continuing for a certain number of cycles and often enough during that time, (iii) significant harmonic content present in the voltage signal, and (iv) the current signal is recorded as zero or a nominal waveform.

The first criterion is met if the first derivative of the voltage signal exceeds 50\% of nominal peak voltage as given by,
\begin{equation}
    \dfrac{|{V}'(n)|}{\hat{V}_q}>\tau_{\text{F}}
    \label{eq:ferro_deriv}
\end{equation}
where $V'(n)$ is the first derivative of the voltage signal, $\bar{x}_{c}$ is the nominal peak voltage, and $\tau_{\text{F}}=0.5$ is the empirically selected threshold for the minimum ratio of the first derivative of the voltage to the nominal value. The second criterion is met if this threshold is exceeded a minimum of \textbf{five} times, occurs at least every \textbf{three} cycles, and occurs for a length of at least \textbf{five} cycles. The third criterion is met if one of the harmonic currents is greater than \textbf{5\%} of the fundamental. Finally, the fourth criterion is met if the RMS current is recorded as zero or the current signal is nominal, which is characterized by a small number of first derivative sign changes. This nominal condition is given by,
\begin{equation}
   \dfrac{N_{\text{I}}}{N}<\tau_{\text{I}}
    \label{eq:ferro_nominal}
\end{equation}
where $N_{\text{I}}$ is the number of first derivative sign changes in the current as calculated using equation \eqref{eq:deriv_sign_change}, $N$ is the total number of samples in the waveform, and $\tau_{\text{I}}=0.3$ is the empirically selected threshold for the ratio of sign changes to total samples.

\subsubsection{Capacitor Bank Switching}\label{sec:cap_bank_switch}

\begin{figure}[t]
\centerline{\includegraphics[width=\columnwidth]{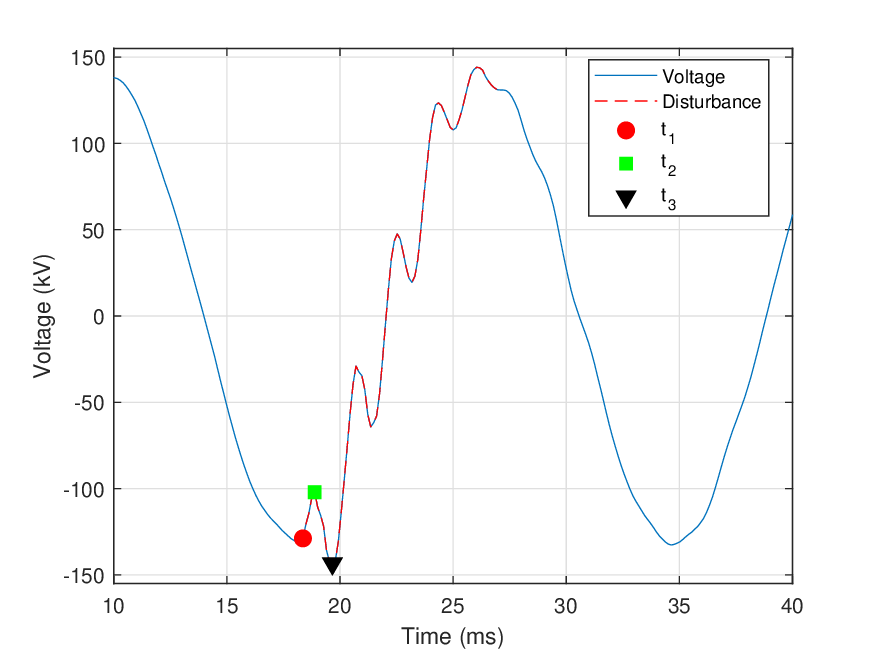}}
\caption{Illustration of a voltage signal collected during capacitor switching event.}
\vspace{-5mm}
\label{fig:cap_switching}
\end{figure}

One of the most common events on a power system 
is capacitor bank switching. Capacitor bank switching induces temporary voltage transients that can create PQ events. 
A typical capacitor bank switching transient is 
characterized by a quick depression of the voltage signal toward zero, 
followed by an overshoot and subsequent transient disturbance--lasting approximately one cycle--as the system returns to 
steady state. These voltage transients may be recorded by devices that are connected to the same bus as the capacitor bank as well as those connected to a different bus. Based upon this fact, the presented automated process is designed to identify capacitor switching for both recording device connection scenarios. 
Fig.~\ref{fig:cap_switching} shows an example of capacitor bank switching in which a broken, red line highlights the portion of the recorded signal associated with the event. 

In a power transmission system, capacitor banks are simultaneously switched in on all three phases. Although Fig.~\ref{fig:cap_switching} shows only a single phase, the other two phase voltage signals are similar in appearance, but will not be identical due the 120$^{\circ}$ phase difference between each of the three signals (i.e., the switching event occurs at different points of the corresponding phase's sinusoidal signal). The disturbance is located within the signal using the first cycle as a reference as described in Sect.~\ref{sec:first_cycle} and shown in Fig.~\ref{fig:disturbance_finder}. The condition for the difference between the actual and ideal voltage signals is given by,
\begin{equation}
    \dfrac{|{V}_{\Delta}|}{\hat{V}_q}>\tau_{\Delta}
    \label{eq:cap_diff}
\end{equation}
where ${V}_{\Delta}$ is the difference between the actual and ideal voltage signals, $\hat{V}_q$ is the nominal peak voltage value, and $\tau_{\Delta}=0.02$ is the threshold empirically selected for this ratio. Once the presence and location of the disturbance has been determined, the disturbance's duration is calculated to ensure that it does not exceed \textbf{two} cycles. 
The voltage signal's peak values must satisfy one of these two criteria: (i) one peak \textbf{2\%} above nominal value and no more than one peak \textbf{10\%} above nominal value; (ii) exactly two peaks \textbf{10\%} above nominal value occurring in neighboring cycles. 

The next step is to determine the three characteristic points highlighted on the waveform of Fig.~\ref{fig:cap_switching}, which are designated as $t_1$ (red circle), $t_2$ (green square), and $t_3$ (black triangle). These points are indicative of a capacitor switching event. First, the portion of the voltage signal one half-cycle before and one half-cycle after the highest peak value is extracted and designated as $V_{\text{O}}$. The point $t_1$ is determined as the first point in which the voltage signal's first derivative exceeded a certain threshold as given by,
\begin{equation}
    \dfrac{|{V}_{\text{O}}'(n)|}{\hat{V}_q}>\tau_{\text{O}}
    \label{eq:cap_deriv}
\end{equation}
where ${V}_{\text{O}}'(n)$ is the first derivative of the overvoltage cycle of the voltage signal, $\hat{V}_q$ is the nominal peak voltage value, and $\tau_{\text{O}}=0.02$ is the threshold empirically selected for this ratio. The first occurrence of this condition is determined to be $t_1$. The point $t_2$ occurs at the lowest point the signal at which the magnitude of voltage signal has dropped below \textbf{90\%} of nominal peak value. The point $t_3$ is then determined as the time index of the highest peak of the voltage signal $V_{\text{O}}$.

The location of these three characteristic points is then validated using the following three checks: (i) the voltage magnitudes at these points are expected the expected values, (ii) nominal number of samples between the overvoltage and the peak prior to it, (iii) the waveform slope is reversed at $t_1$. For the first check, the expected voltage magnitudes at $t_1$, $t_2$, and $t_3$ must follow the inequality given by,
\begin{equation}
    |V_{t_2}|<|V_{t_1}|<|V_{t_3}|, \label{eq:cap_check1}
\end{equation}
where $|V_{t_1}|$, $|V_{t_1}|$, and $|V_{t_3}|$ are the voltage magnitudes at times $t_1$, $t_2$, and $t_3$, respectively. The second check is that the peak before must be approximately equal to $N_c/2$ samples before the overvoltage peak as determined by,
\begin{equation}
    \dfrac{N_{\text{PB}}-N_c/2}{N_c}<\tau_{\text{P}}
    \label{eq:cap_check2}
\end{equation}
where $N_{\text{PB}}$ is the number of samples between the overvoltage peak and the peak before it, $N_c$ is the number of samples in each cycle, and $\tau_{\text{P}}=0.1$ is the threshold empirically selected for this ratio. Finally, the third check was validated using \eqref{eq:deriv_sign_change} described in Sect.~\ref{sec:differentiation}. If the first derivative of the voltage signal leading up to $t_1$ is of opposite sign than the first derivative of the voltage between $t_1$ and $t_2$, then the third check is met. After all these criteria are met for one of the three voltage phases, the other two phases are analyzed to ensure that some form of disturbance is present.

\subsubsection{Lightning Strikes} \label{sec:lightning}
\begin{figure}
\centerline{\includegraphics[width=\columnwidth]{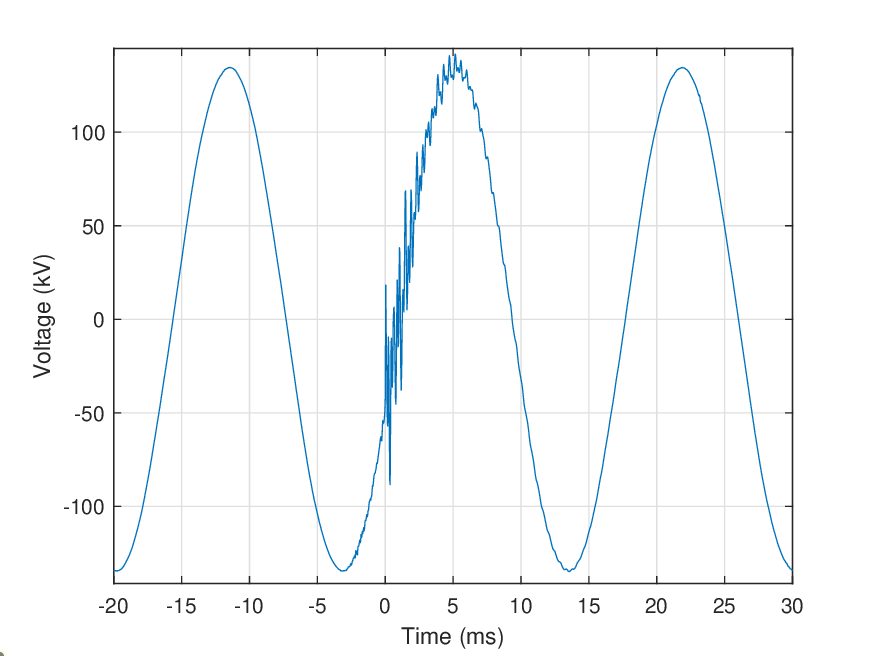}}
\caption{Illustration of a voltage signal collected during a lightning strike event.}
\vspace{-5mm}
\label{fig:lightning}
\end{figure}

Transient overvoltages due to lightning strikes on a transmission line are typically impulses with a rise and decay time in the microseconds. Due to limitations of instrument transformers to pass these high frequencies and 
instrumentation sampling rates, lightning strike events are not readily identified. A representative voltage signal that includes a lightning strike event is shown in Fig.~\ref{fig:lightning}.

First, the automated identification process attempts to identify the event as a capacitor bank switching event (Sect.~\ref{sec:cap_bank_switch}) and then a melting fuse event (Sect.~\ref{sec:fuse_melt}). These steps are taken to ensure that a lightning strike event is not incorrectly identified as either of these two events--that although similar to a lightning strike--are easily distinguished from it as well as one another. If the event is not identified as a capacitor bank switching or melting fuse event, then the disturbance is isolated from the overall signal using the exact same method as that given in equation \eqref{eq:cap_diff} for the isolation of the capacitor bank switching event's disturbance. The disturbance isolation process is repeated for each lightning strike, and the longest strike duration is checked to ensure that it does not exceed \textbf{one} cycle. If more than \textbf{five} disturbances are isolated, then the event is not identified as a lightning strike. In all of the processed data, lightning did not strike more than three times during a single recording. So long as no more than three lightning strike disturbances are isolated, then the automated process identifies the event as a lightning strike and returns the number of strikes along with the disturbance's duration in seconds. 

\subsubsection{Harmonic Resonance}

\begin{figure}[!t]
\centerline{\includegraphics[width=\columnwidth]{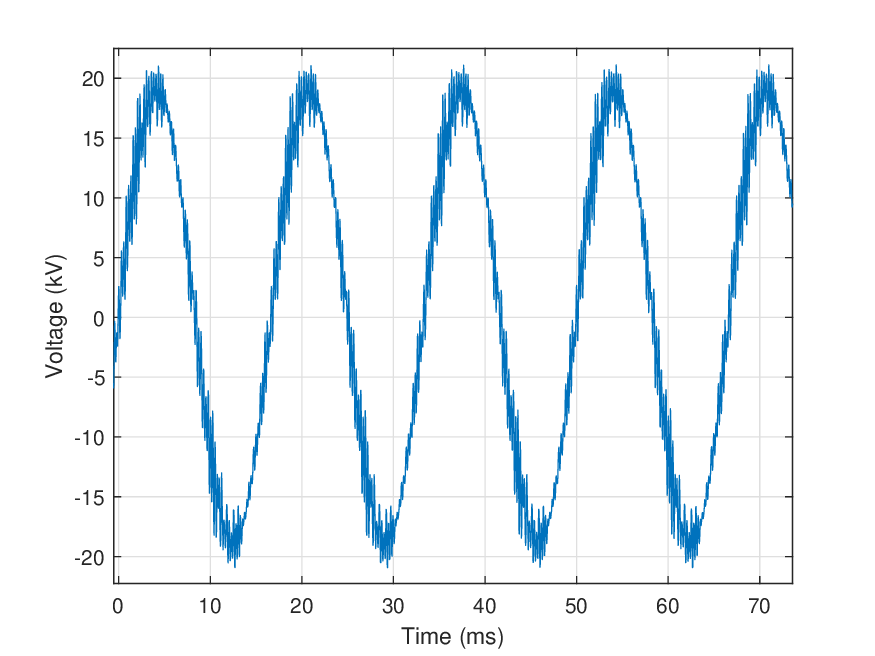}}
\caption{Illustration of a voltage signal collected during a harmonic resonance event.}
\vspace{-5mm}
\label{fig:harm_resonance}
\end{figure}

Power systems have natural frequencies that are a function of the system's inductive and capacitive impedance. 
When a nonlinear load on the power system--such as a VFD--generates a frequency that is 
a natural frequency (i.e., a multiple of the fundamental frequency) of the power system, then a resonance condition can result. 
This resonance can subject equipment to overvoltages or currents, which can 
result in equipment failure or misoperation. Thus, it is important to detect harmonic resonance conditions quickly, so that appropriate and necessary actions can be taken to correct the problem(s).
 Fig.~\ref{fig:harm_resonance} shows an example case of harmonic resonance on an operationally recorded voltage signal.

Harmonic resonance is characterized by the presence of high frequency content in the voltage signals. Based upon this information, the automated identification process first calculates the 
Total Harmonic Distortion (THD) of the voltage signal by, 
\begin{equation}
    V_{\text{THD}}=\dfrac{\sqrt{\sum\limits_{i=2}^{M} |H_i|^{2}}}{H_1},
    \label{eq:thd}
\end{equation}
where $H_i$ is the $i^{\text{th}}$ harmonic, $H_1$ is the fundamental frequency, $M = 100$ is the total number of harmonics used for the calculation, and $|\bullet|$ denotes the magnitude~\cite{bib:thd}. 
If the THD is greater than \textbf{8\%} of the fundamental frequency, then the process continues else it moves onto the next event category. A value of 8\% was empirically selected, but can be adjusted as more data becomes available or based upon power system specifics. 

If the THD threshold is satisfied, then the automated identification process determines whether or not at least the sixth or one of the higher harmonics is more than 5\% of the fundamental frequency's magnitude. If this is the case, then the sign changes in the first derivative are calculated for each cycle using \eqref{eq:deriv_sign_change} as described in Sect.~\ref{sec:differentiation}. The number of first derivative sign changes in each cycle must be at least \textbf{10\%} of the samples in each cycle $N_{c}$ and also occur across \textbf{three} cycles. If all these criteria are satisfied, then the automated process identifies the event as harmonic resonance.

\subsubsection{Improper Voltage Transformer Secondary Grounding}

It is good design practice to use a single and solid grounding point 
on an instrument VT's secondary
~\cite{bib:ieee_grounding}. Otherwise, the result may be incorrect secondary voltage signals in both magnitude and angle, which can lead to the misoperation of protective relays. This can be exacerbated when faults occur on the lines protected by these relays.

\begin{figure}[!t]
\begin{subfigure}{\columnwidth}
    \centering
    \includegraphics[width=\columnwidth]{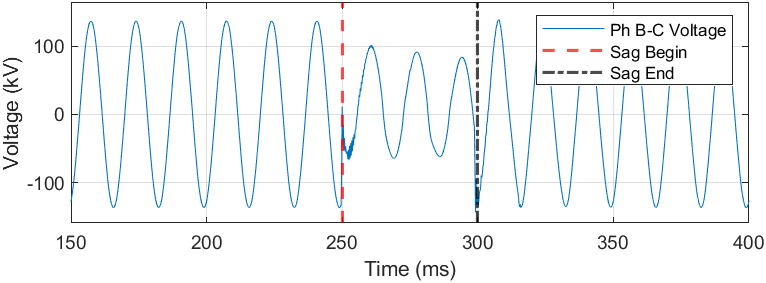}
    \caption{Phase B--C Voltage.}
    \label{fig:vt_ground1}
    \end{subfigure}
    
    \begin{subfigure}{\columnwidth}
    \centering
    \includegraphics[width=\columnwidth]{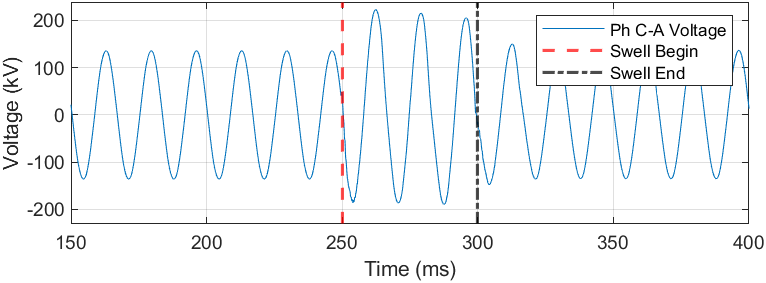}
    \caption{Phase C--A Voltage.}
    \label{fig:vt_ground2}
\end{subfigure}
\centering
\caption{Voltage signals indicating improper VT secondary grounding due to the simultaneous presence of a voltage sag and swell on two phases.}
\vspace{-5mm}
\label{fig:vt_ground}
\end{figure}

A key indicator of improper VT secondary grounding is when one voltage phase is sagged while another one is swelled. Fig.~\ref{fig:vt_ground} provides a representative example of this indicator in which the Phase B--C voltage signal is experiencing a sag from 250~{ms} to 300~{ms} (Fig.~\ref{fig:vt_ground1}) while the Phase C--A voltage signal experiences a swell over the same time period (Fig.~\ref{fig:vt_ground2}). 
Automated identification of improper VT secondary grounding is facilitated by determining if a voltage sag and swell simultaneously exists on two of the three voltage phases. In this work, a sag occurs when one of the voltage signal's peaks \textit{falls} below the nominal peak voltage by more than \textbf{5\%}, and a swell occurs when one of the voltage peaks \textit{rises} above the nominal peak by more than \textbf{5\%}. The phase angle between the sagged and swelled voltage phases is calculated by,
\begin{equation}
    \theta = \cos^{-1}\left(\dfrac{\mathbf{{V_\alpha}}\cdot\mathbf{{V_\beta}}}{|V_\alpha||V_\beta|}\right), 
    \label{dot_product}
\end{equation}
where $\mathbf{V_\alpha}$ and $\mathbf{V_\beta}$ are two of the three faulted voltage phasors, $\cdot$ denotes dot product, and $\theta$ is the phase angle between $\mathbf{V_\alpha}$ and $\mathbf{V_\beta}$. The phase angle is calculated between phases: A to B, B to C, and A to C. In a balanced system, the nominal angle between two voltage phases is 120$^{\circ}$~\cite{bib:power_book}. If the phase angle deviates from this 120$^{\circ}$ nominal angle by more than \textbf{5$^{\circ}$}, then the event is identified as an improper VT secondary grounding event. 

\subsubsection{Incipient Capacitive Voltage Transformer Failure}

\begin{figure}[!t]
\centerline{\includegraphics[width=\columnwidth]{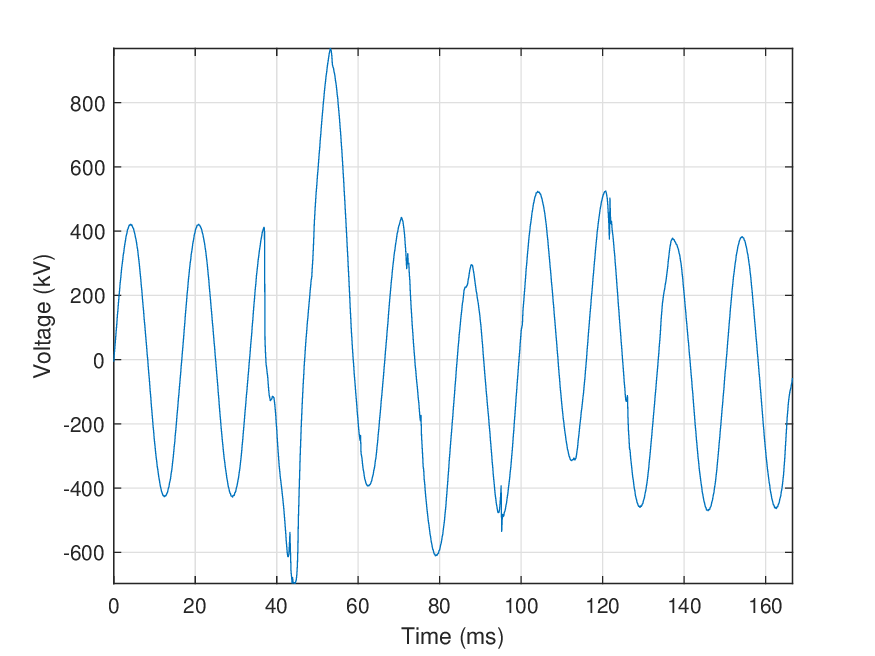}}
\caption{Illustration of a voltage signal showing an incipient CVT failure event.}
\label{fig:cvt}
\end{figure}

Capacitive Voltage Transformers (CVTs) supply voltage to protective relays, so it is very important that the CVT is measuring voltage accurately. 
If 
a catastrophic CVT failure 
results in 
a complete loss of this voltage, then the affected relays detect the loss 
using Loss of Potential (LOP) logic
and act accordingly
~\cite{bib:cvt}.
However, relays are not equipped to detect a CVT that is showing early signs of failure by 
providing incorrect data 
, but has not yet failed to provide the supply voltage. The developed automated identification process is designed to 
detect early indicators of impending CVT failures to facilitate proper actions by utility personnel or equipment. Additionally, a CVT failure poses a significant safety risk to any utility personnel who happen to be nearby when it fails. The voltage signal shown in Fig.~\ref{fig:cvt} provides a representative illustration of the early indicators of an impending (a.k.a., incipient) CVT failure. 

The first indicator of an incipient CVT failure event is that one of the voltage signal's peaks will contain a rise or fall of more than \textbf{10\%} of the nominal peak value, and this behavior must persist for at least \textbf{three} cycles. The second incipient CVT failure indicator is that the disturbance portion of the voltage signal will differ from its corresponding nominal signal, by more than $\tau_{\Delta}=0.02$ as introduced in Sect.~\ref{sec:first_cycle} and implemented in \eqref{eq:cap_diff}. 
Since CVTs are single-phase devices, incipient CVT failure would also only occur in one phase, which is a differentiating factor between other events. Finally, the current signal is analyzed to ensure that no disturbance is present since this event type is specific to voltage signals.

\section{Results}\label{sec:results}

The performance of the developed rule-based, automated electrical disturbance identification process is assessed using a data set comprised of 160 total event records that were collected by field devices operating in a power utility's transmission system. This data set contains approximately ten records for each of the discussed events. The data set also contains events with undisturbed voltage and current signals as well as single--phase and multi--phase events. Each phase of every single--phase event is processed, thus tripling the size of the associated event's data set. False positive and false negative event identifications are counted as incorrect or mis-identifications. 
If a signal did not contain one of the listed electrical disturbances and the automated process did not identify it as a disturbance, then it 
was counted as a correct result. Overall automated identification results are presented in Table~\ref{table:results} for each of the fourteen event types. Table~\ref{table:results} provides the: number of events analyzed, number correctly identified, and the percent correct accuracy for each event type.

\begin{table}[!t]
\centering
\caption{Automated electrical disturbance event identification performance results.
}
\label{table:results}
\begin{tabular}{lllll}
\toprule
\textbf{Event Type}    & \textbf{\# Events} & \textbf{\# Correct} & \textbf{\% Correct} &  \\
\midrule
CT Saturation                & 480          & 464           & 96.67\%    &  \\
\rowcolor{Gray}
A/D Clipping                 & 960          & 953           & 99.27\%    &  \\
Induced Transient Noise      & 480          & 477           & 99.38\%    &  \\
\rowcolor{Gray}
High-Speed Reclosing         & 160          & 160           & 100\%      &  \\
DC Offset                    & 960          & 956           & 99.58\%    &  \\
\rowcolor{Gray}
Motor Starting               & 160          & 160           & 100\%      &  \\
VFD Starting                 & 160          & 160           & 100\%      &  \\
\rowcolor{Gray}
Blown Fuse                   & 160          & 159           & 99.38\%    &  \\
Ferroresonance               & 480          & 476           & 99.17\%    &  \\
\rowcolor{Gray}
Capacitor Switching          & 160          & 159           & 99.38\%    &  \\
Lightning                    & 480          & 477           & 99.38\%    &  \\
\rowcolor{Gray}
Harmonic Resonance           & 480          & 480           & 100\%      &  \\
VT Secondary Grounding       & 160          & 159           & 99.38\%    &  \\
\rowcolor{Gray}
CVT Failure                  & 160          & 154           & 96.25\%    &  \\
\bottomrule
\end{tabular}
\end{table}

\subsection{Results: Current Transformer Saturation}

The accuracy of the automated process in determining CT saturation is 96.67\% (i.e., correctly identifying 464 out of 480 total signals processed). The test for CT saturation proved to be challenging due to the complexity of this event. The range of criteria used may not always be met for each CT saturation event. For example, the A/D clipping waveform of Fig.~\ref{fig:ad_clipping} appears to contain CT saturation based on the characteristic ``kneeing'' in the first two cycles of the fault. However, DC offset is not present and the rating of the CT was likely not exceeded, so this event could be incorrectly classified. Also, for most of this testing, a CT ratio of 1,200:5 is used for each event type regardless of the actual CT ratio. This was done for simplicity, but actual CT ratios from COMTRADE configuration files will be used when these tools are implemented in a production environment. When the actual CT ratio is known, then the rated current of the CT will be known and the automated process is 
able to accurately determine whether this rating was exceeded.

\subsection{Results: Analog-to-Digital Converter Clipping}

The accuracy in detecting the A/D converter clipping event is very high achieving an accuracy of 99.27\% (i.e., correctly identifying 953 out of 960 total signals processed). The threshold for the number of consecutive repeated samples is set to four samples. There are some events where clipping looks obvious to the human eye, but the samples that looked repeated are slightly different. Those results are counted as incorrect, even though the automated process functioned properly. Each utility's personnel could decide whether events like these actually are a problem with the A/D converter. The A/D clipping detection methods should return proper results 100\% of the time if the repeated samples have the exact same value. If they do not, then a very small tolerance (e.g., 10~{V} or 1~{A}) could be allowed between the magnitudes of samples that appear to be the same value.

\subsection{Results: Induced Transient Noise from Switching}

Initial identification performance for this event was poor at roughly 70\%. In an effort to improve automated identification of induced transient noise from switching events, the automated process was modified by incorporating a rule in which the presence of ferroresonance is checked first, then harmonic resonance, and finally induced transient noise from switching so that the three events not take place at the same time. The reason for this is purely due to the similarity with other events and the lack of distinguishing characteristics in this one. Also, a change was made to use the first cycle as a reference to isolate the disturbance as described in Sect.~\ref{sec:first_cycle}. These changes result in an improved accuracy of 99.38\% (i.e., correctly identifying 477 out of 480 total signals processed).

\subsection{Results: High-Speed Reclosing with Tapped Motor Loads}

The accuracy of this event was 100\% in the tests that were conducted. However, there were only two events in which the voltage did not sufficiently decay before reclosing since these events do not often occur if utilities are aware of special settings that are needed for reclosers on such lines with tapped motor loads. Thus, a larger data set will be needed to determine the accuracy of this algorithm.

\subsection{Results: DC Offset}

The DC offset algorithm is one that is well-suited for rule-based analytics as shown by its accuracy of 99.58\% (i.e., correctly identifying 956 out of 960 total signals processed). The frequency analysis method combined with the cycle mean method are very accurate at identifying DC offset. A few signals were falsely classified as DC offset. Signals such as the CT saturation example in Fig.~\ref{fig:ct_sat} contain a very steep spike at the fault inception, so DC offset will be seen in that first half-cycle. Further logic could be added in future work to account for these faults so that DC offset is not detected in the first half-cycle.

\subsection{Results: Motor Starting}

Motor starting events were very straightforward to identify. 160 out of 160 total signals were correctly identified. One reason for the 100\% accuracy is that the other events analyzed did not have many similarities with motor starting. Transformer inrush would produce a similar signal signature, but the differentiating factor is that motor starting is not as rich in harmonics. Motor inrush is also different from single-phase (i.e., the most often occurring) faults in that the elevated current always occurs across all three phases. For these reasons, the motor inrush classification process should be one of the most robust.

\subsection{Results: Variable Frequency Drive Motor Starting}

This event type also produced a 100\% accuracy when tested (i.e., correctly identifying 160 out of 160 total signals processed). However, the 10 VFD starting events used were all from the same motor on the transmission system since these devices are not extremely common. More data will be needed to test the accuracy of the process for this event type.

\subsection{Results: Melted Fuse}

The accuracy in classifying melted fuse events is 99.38\% as it correctly identified 159 out of 159 total signals. Melted fuse events are relatively straightforward to identify due to their short duration. One incorrect classification stemmed from an event containing a minor fault that was incorrectly labeled as a fuse fault. Although the fault lasted several cycles, the part of the current that exceeded the threshold was short enough to be classified as a blown fuse. The process of finding the fault inception and clearing points is very nuanced, and it may not always be 100\% accurate in determining the clearing time, especially for faults that do not greatly (e.g., two times the rated current) exceed the predefined threshold.

\subsection{Results: Ferroresonance}

Ferroresonance is a unique event that was classified with 99.17\% accuracy by these analytics (i.e., correctly identifying 476 out of 480 total signals processed). In most of the data studied, the signals contain large gaps between samples (i.e., at least 50\% of nominal peak value). A few signals did not have such large gaps, possibly due to the ferroresonance being less severe. These events were not identified as ferroresonance, so new methods will need to be developed in the identification of these events. One such method could be incorporating breaker statuses (i.e., open or closed) into the process since ferroresonance usually occurs with the breaker(s) in the open state.

\subsection{Results: Capacitor Switching}

The capacitor switching classification process correctly identified 159 out of 160 total signals resulting in an accuracy of 99.38\%. The methods employed for this event type are very detailed and are much more likely to generate false negatives than false positives. As long as the characteristic three points on the signal of Fig.~\ref{fig:cap_switching} are present, the results should be accurate. The only capacitor switching event that was missed was one in which the voltage transient occurred on the first cycle. Since the nominal peak value is taken using the first cycle as a reference, the rest of the processing becomes incorrect. This issue could be solved by using a predefined nominal peak value for each voltage level from an external data source.

\begin{figure}[!t]
\centerline{\includegraphics[width=\columnwidth]{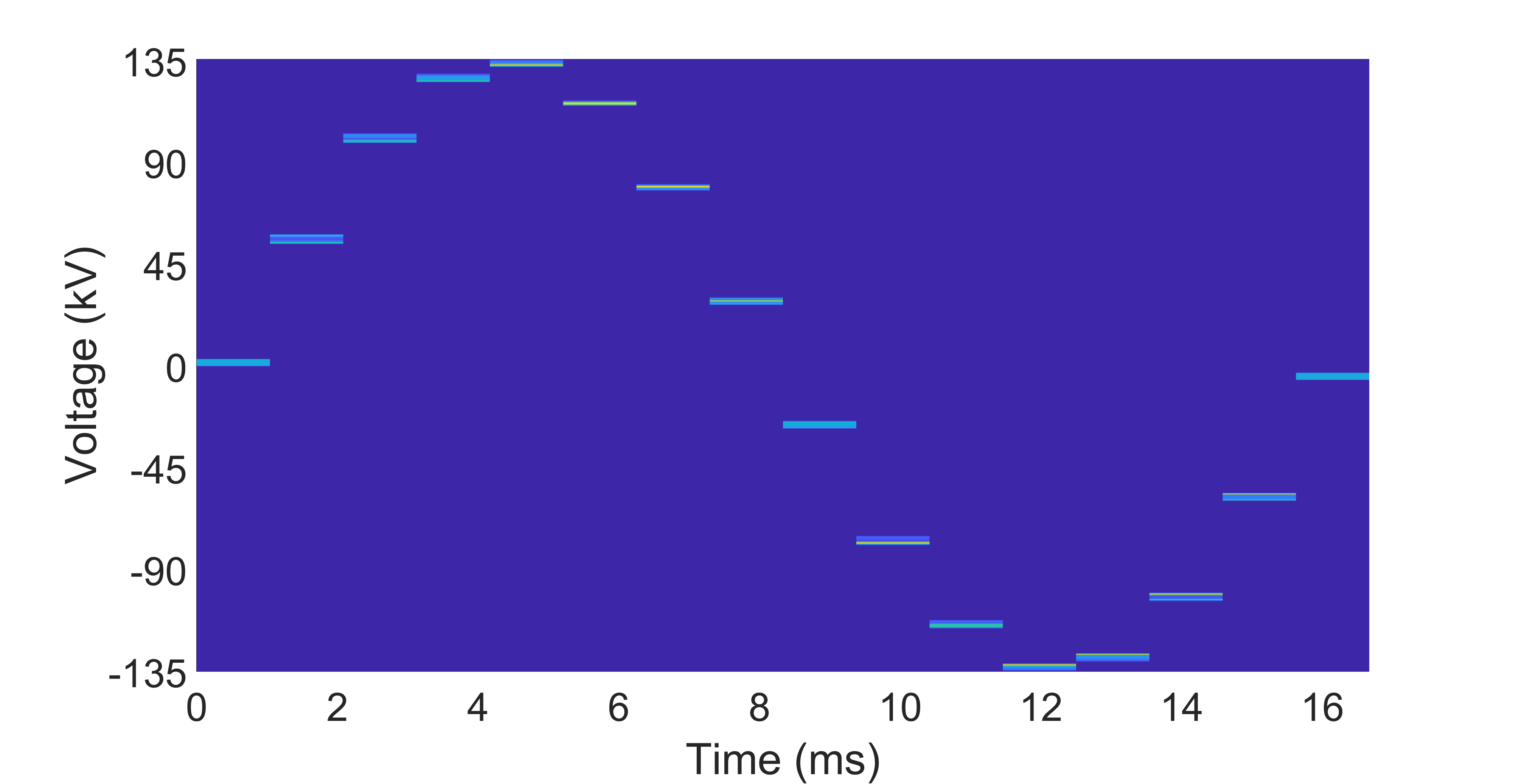}}
\caption{Voltage magnitude cyclic histogram for one hour of operational data from a 161~{kV} transformer DFR.}
\vspace{-5mm}
\label{fig:cyclic_hist}
\end{figure}

\subsection{Results: Lightning}

The automated process correctly identified whether or not lightning was present for 477 out of 480 signals for an accuracy of 99.38\%. Originally, many capacitor switching events were characterized as lightning. To remedy this, the process was updated such that the presence of lightning would only be checked if capacitor switching returned negative. The lightning detection process relies on an accurate determination of the duration of the disturbance. A short disturbance distinguishes lightning from other events. A few events were discovered in which the algorithm determined the disturbance to be longer than it was, which could possibly be due to an outside disturbance unrelated to lightning. This phenomenon results in a few mis-classifications.

\subsection{Results: Harmonic Resonance}

Harmonic resonance is difficult to distinguish from ferroresonance, so a modification was made to only run the harmonic resonance algorithm if ferroresonance has not occurred. This resulted in an accuracy of 100\% with 480 out of 480 signals being correctly identified. There are 5 different harmonic resonance event records in the data set for a total of 15 voltage signals, so more data will be needed to test the robustness of this algorithm. One future improvement that could be made is to detect resonance under the 5\textsuperscript{th} harmonic since resonance conditions can sometimes develop that those frequencies.

\subsection{Results: Voltage Transformer Secondary Grounding}

The classification for this event was very successful with an accuracy of 99.38\% on the 160 signals studied. There are a large number of events in which there is improper VT secondary grounding. Many of the CT saturation faults are not exactly 120$^\circ$ apart in their voltage phase angles, which would indicate improper grounding. This event is straightforward to classify by rule-based techniques. The only issue that may occur is if inaccurate data is fed into the automated process.

\subsection{Results: Incipient Capacitive Voltage Transformer Failure}

The results for this event are not as accurate as the others studied. 154 out of 160 signals (i.e., 96.25\%) were correctly classified as demonstrating incipient CVT failure or not. The lower accuracy is due to the inconsistency in CVT failure events. CVTs could be in different stages of their incipient failure, so the signal signatures will not look the same. The differentiating factor though is that these events are assumed to only occur on one phase at a time, which improves the results.

\begin{figure}[!t]
\centerline{\includegraphics[width=\columnwidth]{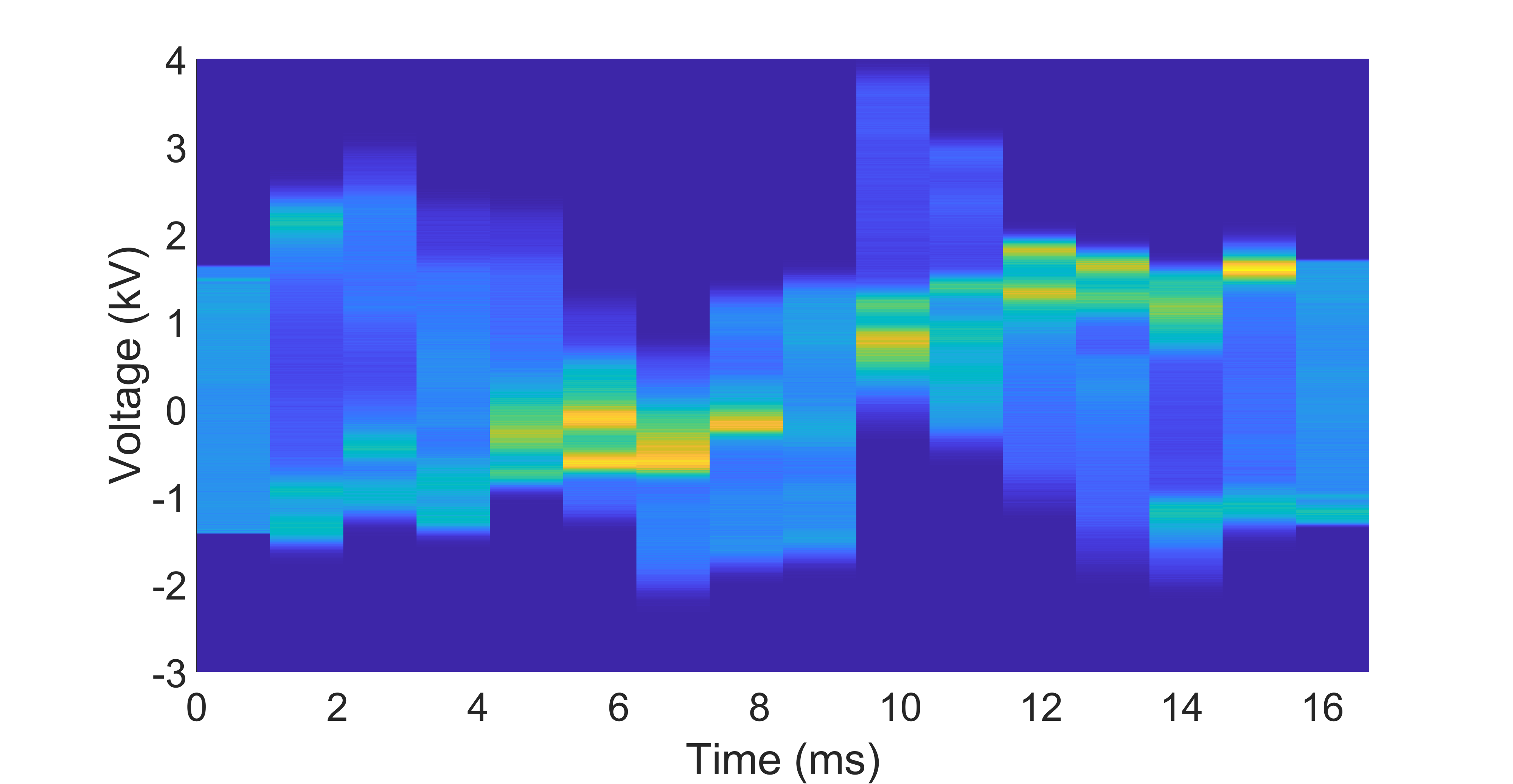}}
\caption{Residual voltage magnitude histogram for one hour of operational data from a 161~{kV} transformer DFR. The range of magnitude has changed from 270~{kV} in the cyclic histogram to 7~{kV} in the residual.}
\vspace{-5mm}
\label{fig:residual_hist}
\end{figure}

\begin{figure}[!b]\vspace{-5mm}
\centerline{\includegraphics[width=\columnwidth]{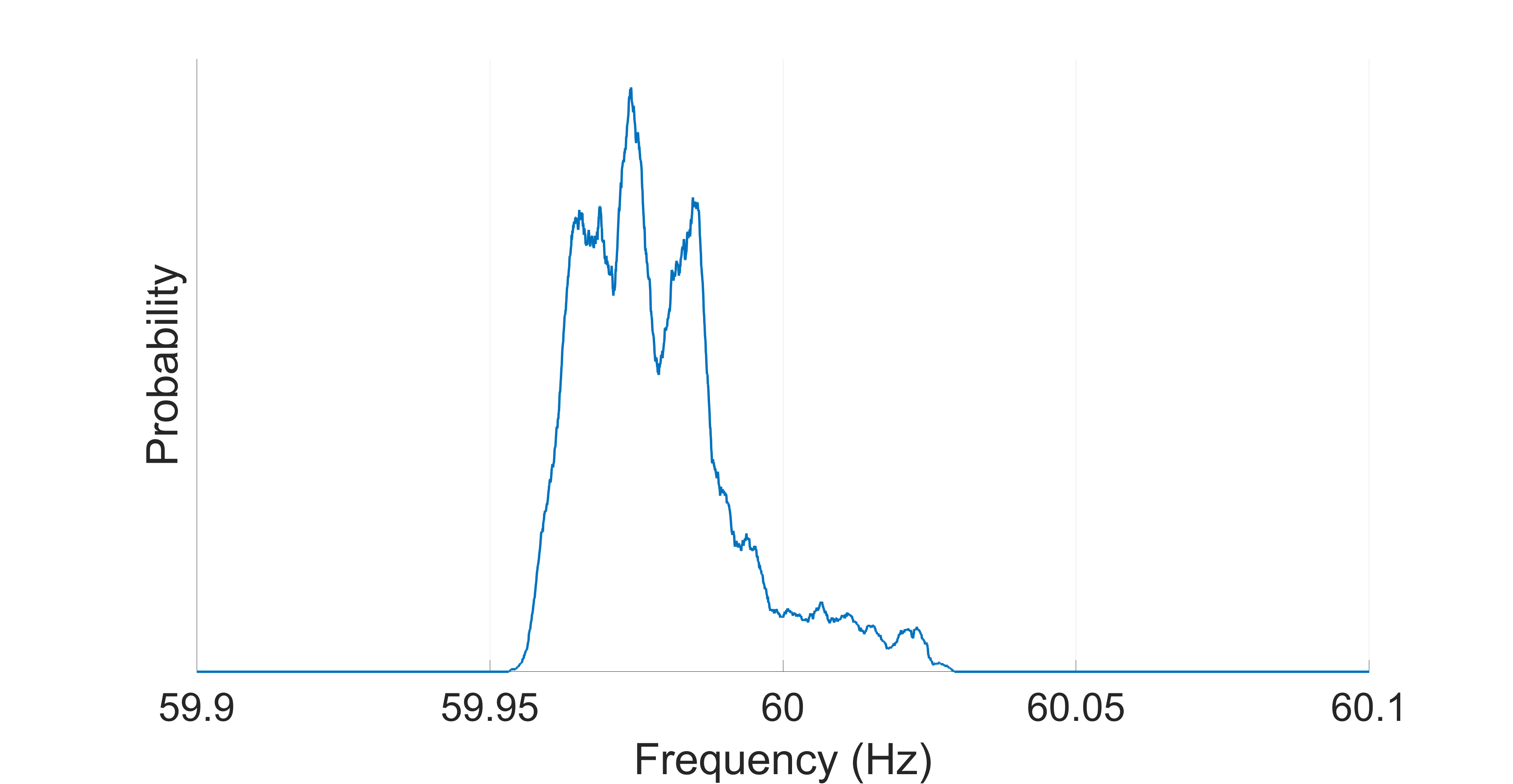}}
\caption{Voltage frequency histogram for one hour of operational data from a 161~{kV} transformer DFR where $f\in[59.9,60.1]$Hz. It is observed that the frequency tends to operate 0.03~{Hz} below the intended 60~{Hz}, but no fault has occurred.}
\label{fig:frequency_hist}
\end{figure}

\subsection{Results: Cyclic Histogram-based Continuous Signal Analysis}

For continuous signal analysis, the developed Python\textsuperscript{\textregistered} script successfully generated the time and frequency-based cyclic histograms and associated residual histograms from a DFR generated CSV file that contains a twenty-four hour period of continuously recorded signals, which includes all three voltage and current signals. After continuous signal processing, the required storage space was reduced by a factor of 320 (i.e., 35~{GB} to 72~{MB}). Fig.~\ref{fig:cyclic_hist} and Fig.~\ref{fig:residual_hist} provide representative examples of the time-based and residual cyclic histograms for one hour, respectively. The same hour of continuous data associated with the cyclic histogram in Fig.~\ref{fig:cyclic_hist} is used to generate the frequency-based histogram in Fig.~\ref{fig:frequency_hist}. Current efforts are focused on integrating the developed Python\textsuperscript{\textregistered} script into a power utility's DFR. Part of this integration involves reducing the amount of DFR compute and memory resources needed to generate the frequency-based histogram and its residual representation. The overarching goal is to use the cyclic histograms to detect deviations within the corresponding signal--that would not ordinarily result in an electrical disturbance event--for incipient prediction, detection, identification, or analysis. Ongoing work is focused on determining the best method of presenting the cyclic histograms so they are informative to PQ engineers. 

\section{Conclusion}\label{sec:conclusion}

In this work, an approach was presented for automated identification of electrical disturbances in a power system. Fourteen different disturbance event types were successfully classified with an average accuracy of 99.13\%, and continuous waveform data was processed and stored using a technique known as a cyclic histogram, which resulted in the file's storage size being reduced in size by a factor of 320. The developed processes will result in time savings for utility personnel as well as increase awareness of disturbances occurring on the power system. This process can categorize events in a matter of minutes rather than hours or days, thus providing utility engineers, operators, and managers with actionable intelligence that will enable immediate and decisive corrective action. Impending--or incipient--device failures will also be detected to enable corrective action before complete failure and so that safety hazards can be removed. This work serves to increase the overall reliability of the transmission system. One goal of future work is to increase the number of disturbance event types that can be classified as well as further test the process using more data. For the continuous waveform analysis portion, future work will involve optimizing the process to reduce computing hardware requirements and further developing the presentation of the data in a useful manner.

\balance

\vspace{12pt}

\begin{IEEEbiography}[{\includegraphics[width=1in,height=1.25in,clip,keepaspectratio]{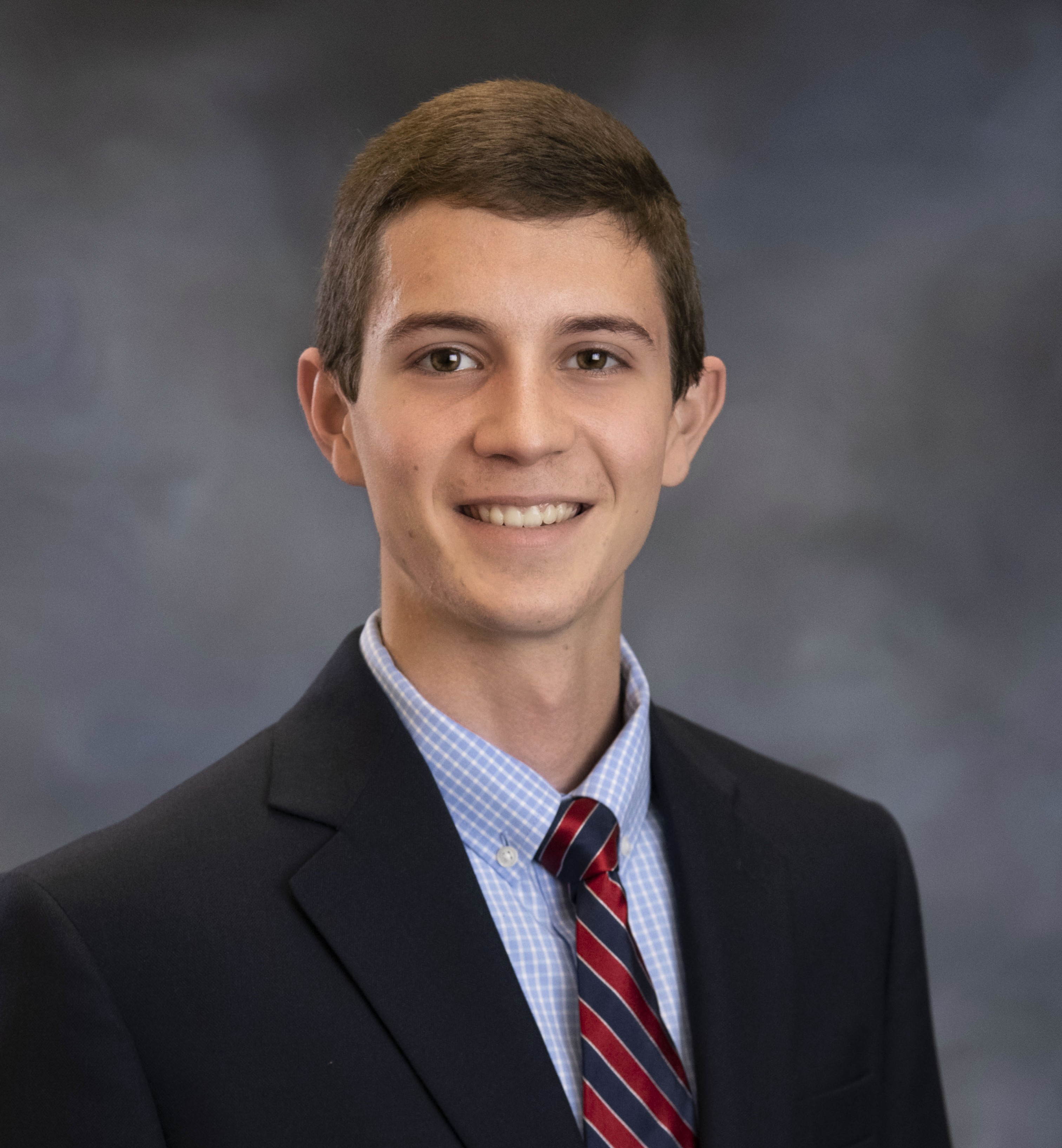}}]{Jonathan D. Boyd} is a master's degree student and researcher at the University of Tennessee at Chattanooga. His research interests include automated power quality event identification and voltage unbalance prediction using machine learning techniques. He graduated with a Bachelor of Science in Electrical Engineering from the University of Tennessee at Chattanooga in 2021. He is currently employed by the Tennessee Valley Authority in the SCADA \& Reliability Systems group.
\end{IEEEbiography}

\begin{IEEEbiography}[{\includegraphics[width=1in,height=1.25in,clip,keepaspectratio]{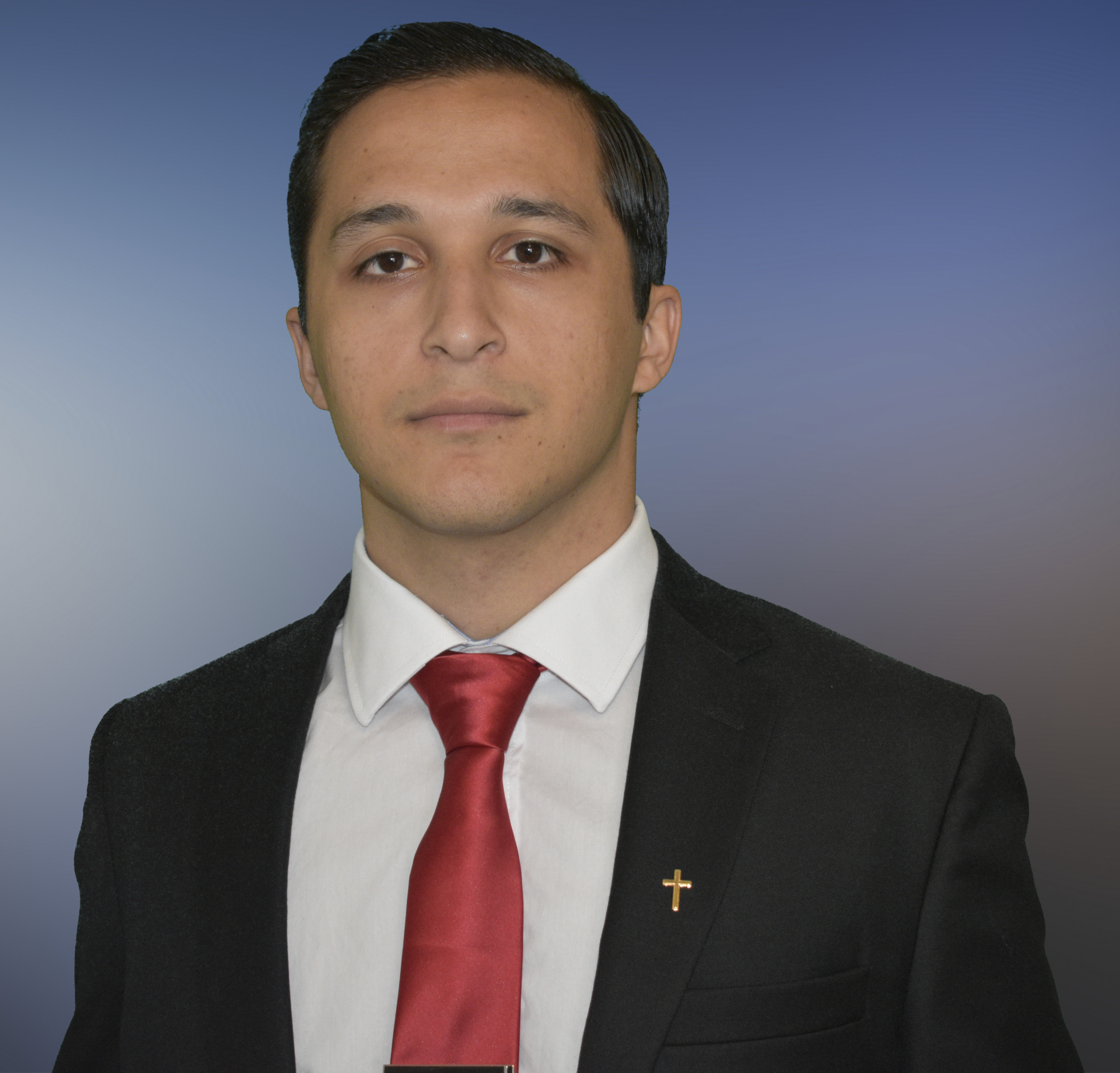}}]{Joshua H. Tyler} is a master's degree candidate and researcher at the University of Tennessee at Chattanooga. His research interests include specific emitter identification, power quality analysis, digital communications, and deep learning. He graduated with a Bachelor of Science in Electrical Engineering from the University of Tennessee at Chattanooga in 2020. He is currently employed by UTC as a research assistant in the Electrical Engineering department.
\end{IEEEbiography}

\begin{IEEEbiography}[{\includegraphics[width=1in,height=1.25in,clip,keepaspectratio]{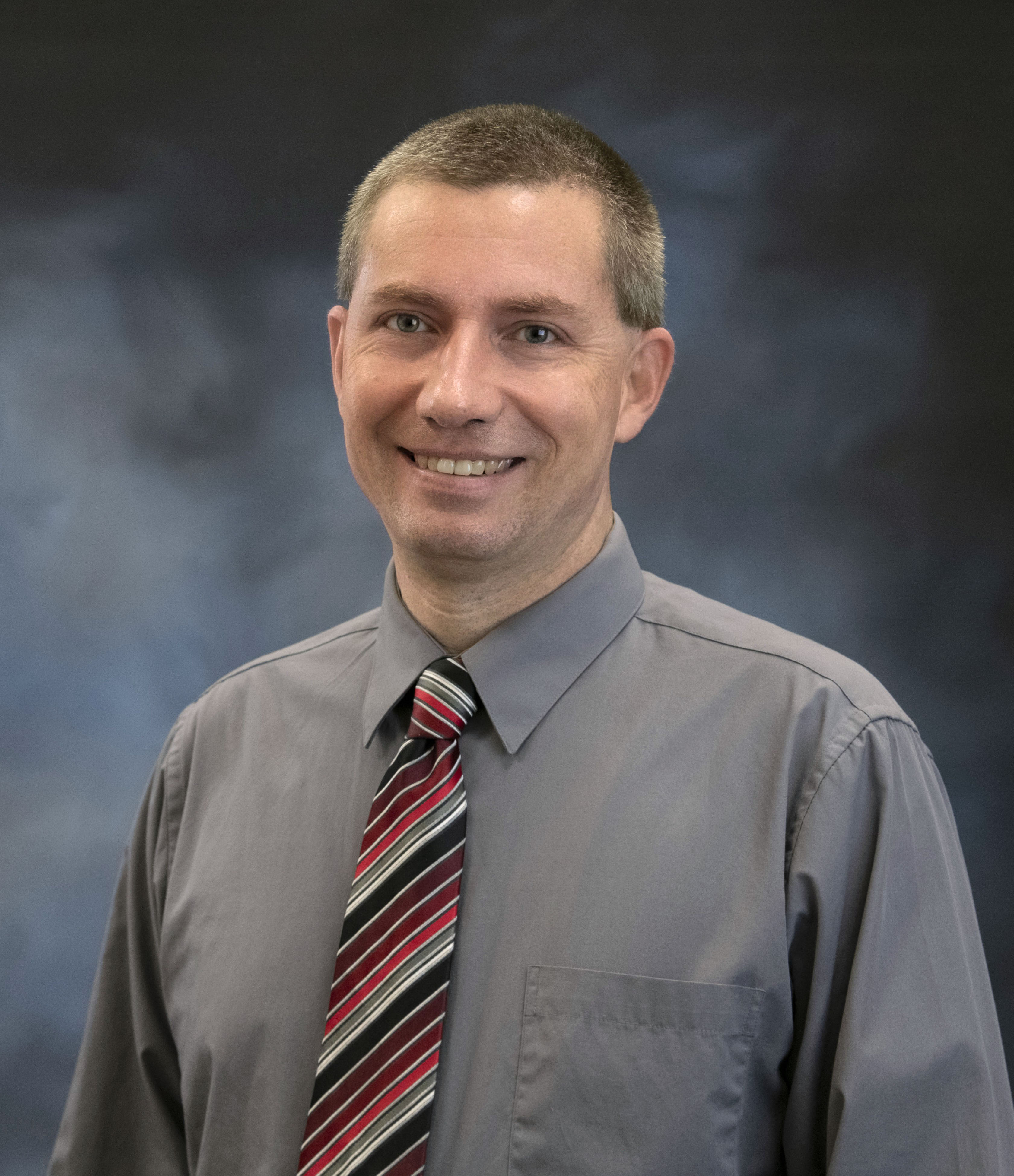}}]{Anthony M. Murphy} currently serves as Senior Program Manager of Power Quality in the Transmission Operations organization at the Tennessee Valley Authority in Chattanooga, TN. He has been employed at TVA for 24 years. In his current role, he works to prevent and address power quality issues impacting the TVA system and TVA customers. Anthony is a registered professional engineer and holds a Bachelor of Science in Electrical Engineering from the University of Tennessee at Knoxville.
\end{IEEEbiography}

\begin{IEEEbiography}[{\includegraphics[width=1in,height=1.25in,clip,keepaspectratio]{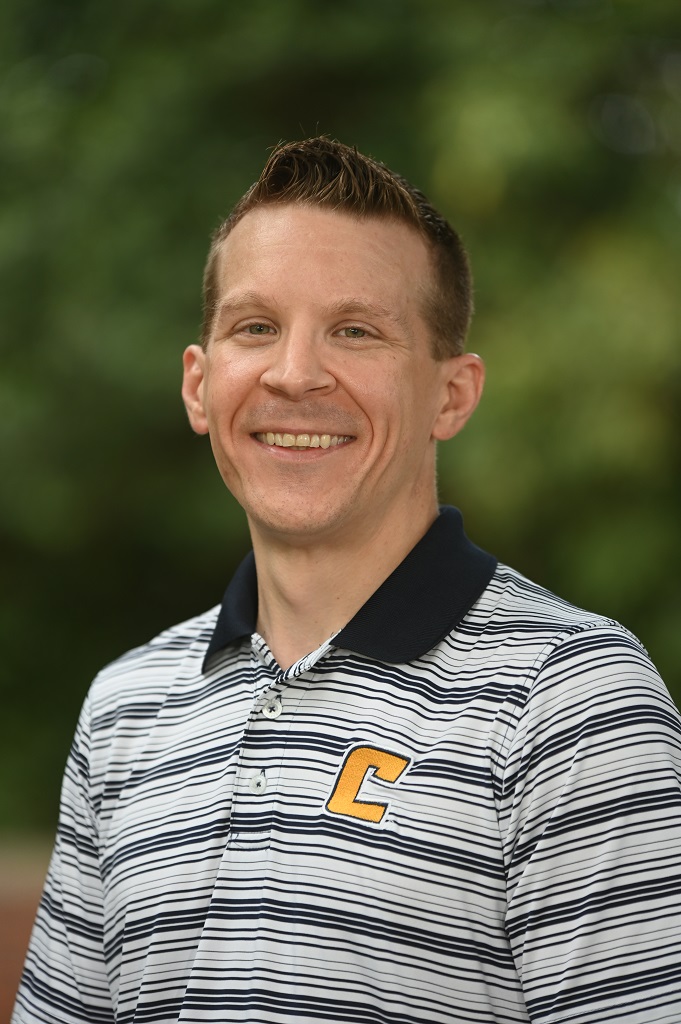}}]{Donald R. Reising} received the B.S. degree in electrical engineering from the University of Cincinnati, Cincinnati, OH in 2006, and the M.S. and Ph.D. degrees in electrical engineering from the Air Force Institute of Technology, Dayton, OH in 2009 and 2012, respectively. From 2008 to 2014, he was an electronics engineer with the U.S. Air Force Research Laboratory in Wright-Patterson Air Force Base, Dayton, OH. He is currently an Associate Professor of Electrical Engineering with the University of Tennessee at Chattanooga. His research interests include wireless device discrimination using RF distinct native attribute fingerprints, compressive sensing, cognitive radio, and deep learning. He is a member of Eta Kappa Nu and Tau Beta Pi.
\end{IEEEbiography}

\end{document}